\documentclass[preprint]{elsarticle}

\usepackage{lineno,hyperref}
\modulolinenumbers[5]

\usepackage[utf8]{inputenc}
\inputencoding{utf8}

\newcommand{\BibTeX}{\textsc{B\kern-0.1emi\kern-0.017emb}\kern-0.15em\TeX}

\usepackage{amssymb, amsmath, amsthm, bm, bbold}
\usepackage{enumitem}
\usepackage{subcaption}
\usepackage{algorithm, algpseudocode}
\usepackage{array, colortbl}
\usepackage{makecell}

\usepackage{cleveref}

\newtheorem{lemma}{Lemma}

\newcommand{\nobs}{n}

\newcommand{\nnu}{\nobs + \nu}

\newcommand{\bfzero}{\mathbf{0}} 
\newcommand{\cov}{\mathbf{\Sigma}} 
\newcommand{\cove}{\Sigma} 
\newcommand{\icov}{\mathbf{\Omega}} 
\newcommand{\icove}{\Omega} 
\newcommand{\scm}{\mathbf{S}} 
\newcommand{\scme}{S} 
\newcommand{\cor}{\mathbf{R}} 
\newcommand{\core}{R} 
\newcommand{\scor}{\mathbf{C}} 
\newcommand{\score}{C} 
\newcommand{\scl}{\mathbf{\Psi}} 
\newcommand{\scle}{\Psi} 
\newcommand{\var}{{\textit{X}}} 
\newcommand{\varv}{{\mathbf{x}}} 
\newcommand{\data}{{\mathbf{D}}} 
\newcommand{\model}{{\mathcal{M}}} 
\newcommand{\given}{\:|\:} 
\newcommand{\Bf}{\mathcal{B}} 
\newcommand{\idm}{\mathbf{I}} 
\newcommand{\imod}{{\mathcal{I}}} 

\newcommand{\unif}{\mathcal{U}}
\newcommand{\Gauss}{\mathcal{N}} 
\newcommand{\Wish}[1]{\mathcal{W}_{#1}} 
\newcommand{\iWish}[1]{\Wish{#1}^{-1}} 
\newcommand{\pWish}{{\mathcal{PW}}} 
\newcommand{\ipWish}{{\pWish^{-1}}} 

\newcommand{\tr}[1]{\textrm{tr}\left(#1\right)}
\newcommand{\diff}[1]{\:\textrm{d}{#1}\:}
\newcommand{\half}[1]{{\frac{#1}{2}}}
\newcommand{\trp}{\intercal} 
\newcommand{\bigO}[1]{\mathcal{O}(#1)}

\newcolumntype{C}[1]{>{\centering\arraybackslash}m{#1}}

\makeatletter
\newcommand*{\indep}{%
	\mathbin{%
		\mathpalette{\@indep}{}%
	}%
}
\newcommand*{\nindep}{%
	\mathbin{
		\mathpalette{\@indep}{\not}
	}%
}
\newcommand*{\@indep}[2]{%
	\sbox0{$#1\perp\m@th$}
	\sbox2{$#1=$}
	\sbox4{$#1\vcenter{}$}
	\rlap{\copy0}
	\dimen@=\dimexpr\ht2-\ht4-.2pt\relax
	\kern\dimen@
	{#2}%
	\kern\dimen@
	\copy0 
}
\makeatother

\journal{International Journal of Approximate Reasoning}

\begin{document}

\begin{frontmatter}
	
	\title{Large-Scale Local Causal Inference of Gene Regulatory Relationships}
	

	\author[postaladdress]{\corref{correspondingauthor}{Ioan Gabriel Bucur}}
	\cortext[correspondingauthor]{Corresponding author}
	\ead{g.bucur@cs.ru.nl}
	
	\author[postaladdress]{Tom Claassen}
	
	\ead{T.Claassen@science.ru.nl}
	
	\author[postaladdress]{Tom Heskes}
	
	\ead{t.heskes@science.ru.nl}

	\fntext[myfootnote]{© 2019. This manuscript version is made available under the CC-BY-NC-ND 4.0 license \url{https://creativecommons.org/licenses/by-nc-nd/4.0/}.}
	
	\address[postaladdress]{Institute for Computing and Information Sciences, Radboud University, Toernooiveld 212, 6525EC Nijmegen, The Netherlands}
	
	\begin{abstract}
		Gene regulatory networks play a crucial role in controlling an organism's biological processes, which is why there is significant interest in developing computational methods that are able to extract their structure from high-throughput genetic data. Many of these computational methods are designed to infer individual regulatory relationships among genes from data on gene expression. We propose a novel efficient Bayesian method for discovering local causal relationships among triplets of (normally distributed) variables. In our approach, we score covariance structures for each triplet in one go and incorporate available background knowledge in the form of priors to derive posterior probabilities over local causal structures. Our method is flexible in the sense that it allows for different types of causal structures and assumptions. We apply our approach to the task of learning causal regulatory relationships among genes. We show that the proposed algorithm produces stable and conservative posterior probability estimates over local causal structures that can be used to derive an honest ranking of the most meaningful regulatory relationships. We demonstrate the stability and efficacy of our method both on simulated data and on real-world data from an experiment on yeast.
	\end{abstract}
	
	
\end{frontmatter}


\section{Introduction}

Gene regulatory networks (GRNs) play a crucial role in controlling an organism's biological processes, such as cell differentiation and metabolism~\cite{karlebach_modelling_2008}. If we knew the structure of a GRN, we could intervene in the developmental process of the organism, for instance by targeting a specific gene with drugs. In recent years, researchers have developed a number of methods for inferring regulatory relationships from data on gene expression~\cite{chai_review_2014}, the process by which genetic instructions are used to synthesize gene products such as proteins. Gene regulatory relationships are inherently causal: one can manipulate the expression level of one gene (the `cause') to regulate that of another gene (the `effect'). Because of this, many GRN inference algorithms rely on causal modeling.

Causal networks such as GRNs can be inferred globally or locally. In the first approach, one considers causal models over all variables in the modeled system at once and searches globally for the best-fitting causal model. Classic causal discovery methods such as the PC algorithm~\cite{spirtes_causation_2000, spirtes_constructing_2000}, which consists of a series of conditional independence tests on the covariance structure meant to progressively reduce the space
of possible causal models, are designed to learn global causal structures. Zhang et al.~\cite{zhang_inferring_2012} apply the PC algorithm to the task of GRN inference by using the conditional mutual information for determining conditional independence relationships. Alternatively, Wehrli and Husmeier~\cite{werhli_reconstructing_2007} attempt to reconstruct gene regulatory networks by combining expression data with multiple sources of biological prior knowledge and sampling over Bayesian network structures. 

Local learning of gene regulatory networks is targeted at identifying a subset of causal regulatory relationships or even a single relationship. The advantage of this second approach lies in its scalability and ease of interpretation~\cite{silverstein_scalable_2000}. The disadvantage is that the global network cannot be easily reconstructed from the (possibly contradicting) inferred local relationships. Yoo and Cooper \cite{yoo_discovery_2002} and later Yoo~\cite{yoo_bayesian_2012} describe Bayesian methods for inferring local causal structures from a combination of observational and experimental data. Luo et al.~\cite{luo_learning_2008}, on the other hand, suggest an approach that only requires observational data, which is based on computing the three-way mutual information of variable triplets.

An efficient way to derive causal relationships from observational data, which results in clear and easily interpretable output, is to find local causal structure in the data. The \textit{local causal discovery} (LCD) algorithm~\citep{cooper_simple_1997} makes use of a combination of observational data and background knowledge when searching for unconfounded causal relationships among triplets of variables. The `Trigger' algorithm~\cite{chen_harnessing_2007} is designed to search for this LCD pattern in gene expression data, using the background knowledge that genetic information is randomized at birth, before any other measurements can be made. In a similar vein, Millstein et al.~\cite{millstein_disentangling_2009} and Neto et al.~\cite{neto_modeling_2013} develop model selection tests for distinguishing among local causal structures over variable triplets. Mani et al.~\citep{mani_theoretical_2006}, on the other hand, divide the causal discovery task into identifying so-called Y structures on subsets of four variables. The Y structure is the smallest structure containing an unconfounded causal relationship that can be learned solely from observational data in the presence of latent variables. 

A key feature of Trigger is that it can estimate the probability of causal regulatory relationships, while controlling for the false discovery rate~\citep{chen_harnessing_2007}. The algorithm consists of a series of likelihood ratio tests for regression coefficients that are translated into statements about conditional (in)dependence, which are then used to identify the presence of the LCD pattern. Testing whether regression coefficients are significantly different from zero essentially boils down to testing whether partial correlation coefficients are significantly different from zero~\citep{meinshausen_high-dimensional_2006}, which means that all the information needed for the tests lies in the covariance structure.

We propose a Bayesian approach for local causal discovery on triplets of (normally distributed) variables that makes use of the information in the covariance structure. With our method, we do not aim to obtain a full reconstruction of the complex gene regulatory networks, but instead focus on the more feasible goal of finding the most promising causal regulatory relationships. To achieve this, we directly score all possible three-dimensional covariance structures in one go and then derive posterior probabilities over local causal structures, with the end goal of identifying plausible causal relationships. This provides a stable, efficient and elegant way of expressing the uncertainty in the underlying local causal structure, even in the presence of latent variables. Moreover, it is straightforward to incorporate background knowledge in the form of priors on causal structures. We show how we can plug in our method into an algorithm that searches for local causal structures in a GRN and outputs a well-calibrated and reliable ranking of the most likely causal regulatory relationships.

The rest of the paper is organized as follows. In Section~\ref{sec:bkg}, we introduce some standard background notation and terminology. In Section~\ref{sec:bfcs}, we describe our Bayesian approach for inferring the covariance structure of a three-dimensional Gaussian random vector. By defining simple priors, we then derive the posterior probabilities of local causal structures given the data. In Section~\ref{sec:experiments}, we apply and evaluate our method on simulated and real-world data. In Section~\ref{sec:complexity}, we then analyze its computational and time complexity. We conclude by discussing advantages and disadvantages of our approach in Section~\ref{sec:discussion}.

\section{Background} \label{sec:bkg}

Causal structures can be represented by directed graphs, where the nodes in the graph represent (random) variables and the edges between nodes represent causal relationships. \textit{Maximal ancestral graphs} (MAGs) encode conditional independence information and causal relationships in the presence of latent variables and selection bias~\citep{richardson_ancestral_2002}. We refer to MAGs without undirected edges as \textit{directed maximal ancestral graphs} (DMAGs). DMAGs are closed under marginalization, which means they preserve the conditional independence information in the presence of latent variables.

Two causal structures are \textit{Markov (independence) equivalent} if they imply the same conditional independence statements. The \textit{Markov equivalence class} of a MAG (or DMAG) is represented by a \text{partial ancestral graph} (PAG), which displays all the edge marks (arrowhead or tail) shared by all members in the class and displays circles for those marks that are not common among all members. In this work, we will consider two types of graphs: \textit{directed acyclic graphs} and \textit{directed maximal ancestral graphs}. However, the results presented can be applied to any causal graph structure.

A \textit{(conditional) independence model} $\imod$ (\textit{CI model} for short) over a finite set of variables $V$ is a set of triples $\left<X, Y \given Z\right>$, called \textit{(conditional) independence statements}, where $X, Y, Z$ are disjoint subsets of $V$ and $Z$ may be empty~\citep{studeny_probabilistic_2006}. We can induce a (probabilistic) independence model over a probability distribution $P \in \mathcal{P}$ by letting:

$$\left<A, B \given C\right> \in \imod(P) \iff A \indep B \given C \textrm{ w.r.t. } P.$$

The conditional independence model induced by a multivariate Gaussian distribution is a \textit{compositional graphoid}~\citep{sadeghi_markov_2014}, which means that it satisfies the \textit{graphoid axioms} and the \textit{composition} property. Because of this, there is a one-to-one correspondence between the conditional independence models that can be induced by a multivariate Gaussian and the Markov equivalence classes of a causal graph structure.

\section{Bayes Factors of Covariance Structures (BFCS)}  \label{sec:bfcs}

\begin{figure}[!htb]
	\centering
	\begin{tabular}{c | C{.46\linewidth} | C{.12\linewidth} C{.12\linewidth}}
		CI Model & Markov Equivalence Class (PAG) & Covariance Matrix & Precision Matrix \\
		\hline \hline
		&&& \\
		\makecell{`$\var_1 \nindep \var_2 \nindep \var_3$' \\ (Full)} &
		\includegraphics{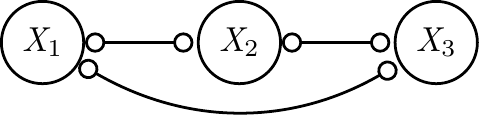} &
		\includegraphics{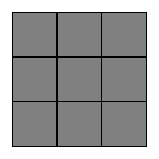} &
		\includegraphics{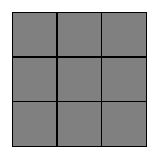} \\
		\makecell{`$\var_1 \indep \var_3$' \\ (Acausal)} &
		\includegraphics{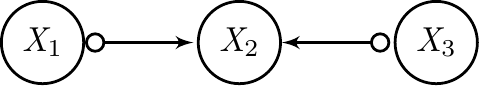} &
		\includegraphics{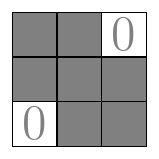} &
		\includegraphics{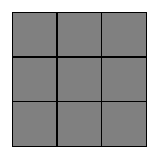} \\
		\makecell{`$\var_1 \indep \var_3 \given \var_2$' \\ (Causal)} & 
		\includegraphics{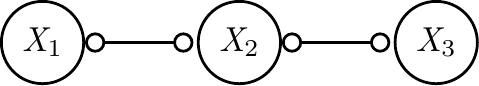} &
		\includegraphics{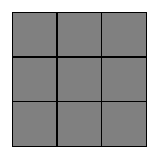} &
		\includegraphics{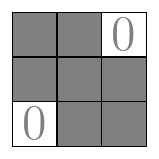} \\
		\makecell{`$(\var_1, \var_3) \indep \var_2$' \\ (Independent)} &
		\includegraphics{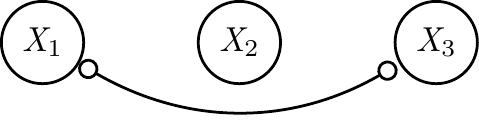} &
		\includegraphics{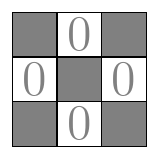} &
		\includegraphics{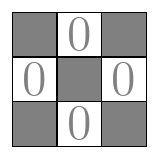} \\
		\makecell{`$\var_1 \indep \var_2 \indep \var_3$' \\ (Empty)} &
		\includegraphics{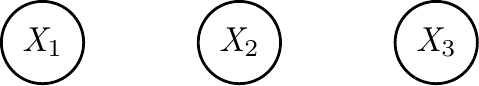} &
		\includegraphics{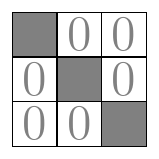} &
		\includegraphics{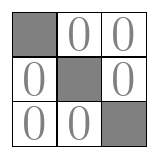} \\
	\end{tabular}
	\caption{Overview of the five canonical cases depicting the equivalence between conditional independence models, Markov equivalence classes and covariance structures.} \label{fig:equivalence}
\end{figure}

We are interested in inferring the local covariance structure from observational data. To arrive at a simple and extremely efficient algorithm, we will be working with triplets of variables and we will assume that the data follows a (latent) Gaussian model. While it is possible to assume another underlying data generation model (given a particular covariance structure), deriving the Bayes factors will in general be much less tractable. With finite data, we can never be sure about the true covariance structure underlying the data. Hence, we prefer to work with probability distributions over covariance matrices. For a general three-dimensional covariance matrix $\cov$, the likelihood reads: 

\begin{equation} \label{eqn:likelihood}
	p(\data \given \cov) = (2\pi)^{-\frac{3n}{2}} |\cov|^{-\frac{n}{2}} \exp\left[-\frac{1}{2} \tr{\scm \cov^{-1}} \right],
\end{equation} 
where $\data$ is the data set containing $\nobs$ independent and identically distributed observations and  $\scm = \data^\trp \data$ is the scatter matrix.

Under the Gaussianity assumption, there is a one-to-one correspondence between the constraints in the covariance matrix and the conditional independences among the variables. There are five specific canonical cases to consider, which are depicted in Figure~\ref{fig:equivalence}. We show in the appendix that these are the only possible canonical cases for non-degenerate (full-rank) covariance structures over three variables (\ref{lem:complete_cov_str}). The `full' and `empty' covariance structures are self-explanatory. We call `independent' the case occurring when one variable is independent of the other two. We call the case on the second row `acausal' because $\var_2$ cannot cause $\var_1$ or $\var_3$ if conditioning upon $\var_2$ turns a conditional independence between $\var_1$ and $\var_3$ into a conditional dependence. We call the case on the third row `causal' because $\var_2$ either causes $\var_1$ or $\var_3$ if conditioning upon $\var_2$ turns a conditional dependence between $\var_1$ and $\var_3$ into a conditional independence~\citep{claassen_logical_2011}. The five cases translate into eleven distinct covariance structures when considering all permutations of three variables. These are the only possible covariance structures on three variables, since the conditional independence model induced by a multivariate Gaussian is a compositional graphoid~\citep{sadeghi_markov_2014}.

Our goal is to compute the posterior probability of each of the possible conditional independence models given the data. We denote by $\mathcal{J} = \{\model_0, \model_1, ..., \model_{10}\}$ the set of all possible conditional independence models. The model evidence (marginal likelihood) is then, for $\model_j \in \mathcal{J}$: $$p(\data \given \model_j) = \int \diff \cov p(\data \given \cov) p(\cov \given \model_j).$$

To facilitate computation, we derive the Bayes factors $\Bf_j$ of each conditional independence model ($\model_j$) compared to a reference model ($\model_0$): 

\begin{equation*}
	\Bf_j = \frac{p(\data \given \model_j)}{p(\data \given \model_0)} = \frac{\int \diff{\cov} p(\data \given \cov) p(\cov \given \model_j)}{\int \diff{\cov} p(\data \given \cov) p(\cov \given \model_0)}.
\end{equation*}

As we shall see in Subsection~\ref{ssec:derivation}, many terms will cancel out, making the resulting ratios much simpler to compute (see for example Equation~\ref{eqn:bf_indep}). Finally, we arrive at the posterior probabilities:

\begin{equation}
p(\model_j \given \data) = \frac{p(\data \given \model_j) \cdot p(\model_j)}{\sum_i p(\data \given \model_i) \cdot p(\model_i)} = \frac{\Bf_j \cdot p(\model_j)}{\sum_i \Bf_i \cdot p(\model_i)},  \label{eqn:post_ci_model}
\end{equation}
where $p(\model_j)$ is the prior probability of the conditional independence model $\model_j \in \mathcal{J}$ and $\Bf_j$ is the Bayes factor of $\model_j$ versus $\model_0$.

\subsection{Choosing the Prior on Covariance Matrices} \label{ssec:choosing_prior_covm}

We consider the inverse Wishart distribution for three-dimensional covariance matrices, denoted by $\iWish{3}$, which is parameterized by the positive definite scale matrix $\scl$ and the number of degrees of freedom $\nu$: 

$$ \cov \sim \iWish{3} (\scl, \nu); \quad  p(\cov) = \frac{|\scl|^{\half{\nu}}}{2^{\half{3 \nu}} \Gamma_3(\half{\nu})}  |\cov|^{-\half{\nu + 4}} \exp\left[ -\half{1} \tr{\scl \cov^{-1}} \right],$$
where $\Gamma_p$ is the $p$-variate gamma function.

The inverse Wishart is the conjugate prior on the covariance matrix of a multivariate Gaussian vector, which means the posterior is also inverse Wishart. Given the data set $\data$ containing $\nobs$ observations and $\scm = \data^\trp \data$ the scatter matrix, the posterior over $\cov$ reads:

\begin{equation} \label{eqn:posterior}
\cov \given \data \sim \iWish{3}(\scl + \scm, \nu + \nobs). \\
\end{equation}

In order to choose appropriate parameters for the inverse Wishart prior, we analyze the implied distribution in the space of correlation matrices. By transforming the covariance matrix into a correlation matrix, we end up with a so-called \textit{projected inverse Wishart} distribution on the latter, which we denote by $\ipWish$. Barnard et al.~\cite{barnard_modeling_2000} have shown that if the correlation matrix $\cor$ follows a projected inverse Wishart distribution with scale parameter $\scl$ and $\nu$ degrees of freedom, then the marginal distribution $p(\core_{ij}), i \ne j,$ for off-diagonal elements is uniform if we take $\scl$ to be any diagonal matrix and $\nu = p + 1$, where $p$ is the number of variables. We are working with three variables, so we choose $\nu = 4$.

It is easy to check that for any diagonal matrix $D$, the projected inverse Wishart is scale invariant: 

$$\ipWish(\scl, \nu) \equiv \ipWish(D \scl D, \nu).$$

From~\eqref{eqn:posterior}, it then follows that we can make the posterior distribution on the correlation matrices independent of the scale of the data by choosing the prior scale matrix $\scl = \bfzero_{3, 3}$, where $\bfzero_{3, 3}$ is the $3 \times 3$ null matrix. Since that would lead to an undefined prior distribution, we can achieve the same goal by setting $\scl = \epsilon \idm_3$ in the limit $\epsilon \downarrow 0$, where $\idm_3$ is the $3 \times 3$ identity matrix and $\epsilon$ is a scalar variable. Summarizing, we will consider the following prior distribution for the covariance matrix: 

\begin{equation} \label{eqn:prior_cov}
\cov \sim \iWish{3}(\epsilon \idm_3, 4), \quad \epsilon \downarrow 0.
\end{equation}

\subsection{Deriving the Bayes Factors} \label{ssec:derivation}

As reference model ($\model_0$), we choose the most general case in which no independences can be found in the data (`$\var_1 \nindep \var_2 \nindep \var_3$'), which means that the covariance matrix is unconstrained (Figure~\ref{fig:equivalence}, first row). We assume that, given $\model_0$ is true, the covariance matrix follows an inverse Wishart distribution

$$p(\cov \given \var_1 \nindep \var_2 \nindep \var_3) = \iWish{3}(\cov; \epsilon \idm_3, \nu),$$ where we consider the limit $\epsilon \downarrow 0$ and set $\nu = 4$ (see Subsection~\ref{ssec:choosing_prior_covm}). Using the conjugacy of the inverse Wishart prior for the likelihood of the covariance matrix in~\eqref{eqn:likelihood}, we immediately get the model evidence

\begin{equation} \label{eqn:ref_evidence}
	p(\data \given \var_1 \nindep \var_2 \nindep \var_3) = \frac{\epsilon^\frac{3\nu}{2} \Gamma_3(\frac{n + \nu}{2})}{\pi^\frac{3n}{2} \Gamma_3(\frac{\nu}{2})} |\scm + \epsilon \idm_3|^{-\frac{n + \nu}{2}},
\end{equation}

where $\Gamma_3$ is the trivariate gamma function, $\data$ is the data set containing $\nobs$ observations, and $\scm = \data^\trp \data$ is the scatter matrix. 

\begin{itemize}
\item We first compare the evidence for the conditional independence model `$\var_1 \indep \var_2 \indep \var_3$' to the evidence for the reference model `$\var_1 \nindep \var_2 \nindep \var_3$' by computing the Bayes factor:

$$ 	\Bf(\var_1 \indep \var_2 \indep \var_3) = \frac{p(\data \given \var_1 \indep \var_2 \indep \var_3)}{p(\data \given \var_1 \nindep \var_2 \nindep \var_3)}.$$

We can implement the `$\var_1 \indep \var_2 \indep \var_3$' case (Figure~\ref{fig:equivalence}, last row) by constraining $\cov$ to be diagonal, which means we only have to consider the parameters $\cove_{11}, \cove_{22}, \cove_{33}$. We propose to take the prior

\begin{equation} \label{eqn:prior_ind}
	p(\cov \given \var_1 \indep \var_2 \indep \var_3) = \prod_{i=1}^3 \iWish{1}(\cove_{ii}; \epsilon, \nu).
\end{equation}

The likelihood in~\eqref{eqn:likelihood} factorizes analogously in this case and becomes

$$ p(\data \given \cov) = \prod_{i=1}^3 \left\{ (2\pi \cove_{ii})^{-\frac{n}{2}} \exp\left[ -\frac{1}{2} \tr{\scme_{ii} \cove^{-1}_{ii}} \right] \right\},$$
yielding the model evidence

$$ p(\data \given \var_1 \indep \var_2 \indep \var_3) =  \prod_{i=1}^3 \left[\frac{\epsilon^\frac{\nu}{2} \Gamma_1(\frac{n + \nu}{2})}{\pi^\frac{\nobs}{2} \Gamma_1(\frac{\nu}{2})} (\scme_{ii} + \epsilon)^{- \frac{\nobs + \nu}{2}}\right].$$
Dividing by the reference model evidence from~\eqref{eqn:ref_evidence} and taking the limit $\epsilon \downarrow 0$, we obtain the Bayes factor

\begin{equation}
	\begin{aligned}
		\Bf(\var_1 \indep \var_2 \indep \var_3) &= \frac{\Gamma_3(\frac{\nu}{2})}{\Gamma_3(\frac{\nobs + \nu}{2})} \left[\frac{\Gamma_1(\frac{\nobs + \nu}{2})}{\Gamma_1(\frac{\nu}{2})}\right]^3 |\scor|^\frac{\nobs + \nu}{2} \\
		&= \frac{\nobs + \nu - 2}{\nu - 2} \frac{\Gamma(\frac{\nu - 1}{2})}{\Gamma(\frac{\nobs + \nu}{2})} \frac{\Gamma(\frac{\nu}{2})}{\Gamma(\frac{\nobs + \nu - 1}{2})} |\scor|^\frac{\nobs + \nu}{2}, \label{eqn:bf_indep}
	\end{aligned}
\end{equation}
with $\scor$ the sample correlation matrix and $\Gamma$ the (univariate) gamma function. Due to the choice~\eqref{eqn:prior_ind}, the evidence for `$\var_1 \indep \var_2 \indep \var_3$' also scales with $\epsilon^{\half{\nu}}$, so the dominant terms depending on $\epsilon$ cancel out and the Bayes factor depends only on the correlation matrix in the limit $\epsilon \downarrow 0$. 

\item We now show how to derive the Bayes factor for the `$\var_3 \indep (\var_1, \var_2)$' case. The derivations for the other two permutations, `$\var_1 \indep (\var_2, \var_3)$' and `$\var_2 \indep (\var_3, \var_1)$', are completely analogous. Given the independence statements, the covariance matrix $\cov$ is block diagonal, consisting of the submatrix $\cov_{(1, 2), (1, 2)}$ (specified by the parameters $\cove_{11}, \cove_{12}, \cove_{22}$) and the single element $\cove_{33}$. We thus propose the following prior: 
	
	$$p(\cov \given \var_3 \indep (\var_1, \var_2)) = \iWish{2}(\cov_{(1, 2), (1, 2)}; \epsilon I_2, \nu) \times \iWish{1}(\cove_{33}; \epsilon, \nu),$$
	The likelihood in~\eqref{eqn:likelihood} factorizes accordingly into two terms: 
	
	\begin{align*}
		p(\data \given \cov) =& (2\pi)^{-\frac{2\nobs}{2}} |\cov_{(1, 2), (1, 2)}|^{-\frac{\nobs}{2}} \exp\left[-\frac{1}{2} \tr{\scm_{(1, 2), (1, 2)}\cov_{(1, 2), (1, 2)}^{-1}}\right] \times \\ 
		& (2\pi)^{-\frac{\nobs}{2}} \cove_{33}^{-\frac{\nobs}{2}} \exp\left[-\frac{1}{2} \scme_{33} \cove_{33}^{-1}\right].
	\end{align*}
	
	The marginal likelihood then reads 
	
	$$p(\data \given \var_3 \indep (\var_1, \var_2)) =  \frac{\epsilon^\frac{3 \nu}{2}}{\pi^\frac{3\nobs}{2}} \frac{\Gamma_1(\frac{\nobs + \nu}{2})}{\Gamma_1(\frac{\nu}{2})} \frac{\Gamma_2(\frac{\nobs + \nu}{2})}{\Gamma_2(\frac{\nu}{2})} (\scme_{33} + \epsilon)^{-\frac{\nobs + \nu}{2}} |\scm_{(1, 2), (1, 2)} + \epsilon I_2|^{-\frac{\nobs + \nu}{2}}.$$
	
	Dividing by~\eqref{eqn:ref_evidence} we obtain the following Bayes factor, in the limit $\epsilon \downarrow 0$:
	
	\begin{align*}
		\Bf(\var_3 \indep (\var_1, \var_2)) &= \frac{\Gamma_3(\half{\nu})}{\Gamma_3(\half{\nnu})} \frac{\Gamma_2(\half{\nnu})}{\Gamma_2(\half{\nu})} \frac{\Gamma_1(\half{\nnu})}{\Gamma_1(\half{\nu})} \left[\frac{|\scor|}{1 - \score_{12}^2}\right]^\half{\nnu} \\
		&= \frac{\nnu - 2}{\nu - 2} \left[\frac{|\scor|}{1 - \score_{12}^2}\right]^\half{\nnu}.
	\end{align*}

\item We now show how to derive the Bayes factor for the `$\var_1 \indep \var_2$' case.  The derivations for the other two permutations, `$\var_2 \indep \var_3$' and `$\var_3 \indep \var_1$', are completely analogous. The inverse-Wishart distribution in~\eqref{eqn:prior_cov} factorizes, for arbitrary $\scl$ and $\nu$, into:
	
	\begin{align*}
		\iWish{3}(\cov; \scl, \nu) = &\iWish{2}(\cov_{(1, 2), (1, 2)}; \scl_{(1, 2), (1, 2)}, \nu - 1) \times \iWish{1}(\cove_{33 \cdot (1, 2)}; \scle_{33 \cdot (1,2)}, \nu) \times \\ & \Gauss(\cov_{(1, 2), (1, 2)}^{-1} \cov_{(1, 2), 3}; \scl_{(1, 2), (1, 2)}^{-1} \scl_{(1, 2), 3}, \cove_{33 \cdot (1, 2)} \scl_{(1, 2), (1, 2)}^{-1}).
	\end{align*}

	We can implement the constraint $\cove_{12} = 0$ by forcing $\cov_{(1, 2), (1, 2)}$ to be diagonal. This suggests the prior:
	\begin{align*}
		p(\cov \given \var_1 \indep \var_2) =& \prod_{i=1}^2 \iWish{1}(\cove_{ii}; \epsilon, \nu - 1) \times \iWish{1}(\cove_{33 \cdot (1, 2)}; \epsilon, \nu) \times \\
		&\Gauss(\cov_{(1, 2), (1, 2)}^{-1} \cov_{(1, 2), 3}; 0, \epsilon^{-1} \cove_{33 \cdot (1, 2)} I_2).
	\end{align*}
	
	Note that in the limit $\epsilon \downarrow 0$, the prior normal component moves towards an improper flat distribution:
	
	$$ \Gauss(\cov_{(1, 2), (1, 2)}^{-1} \cov_{(1, 2), 3}; 0, \epsilon^{-1} \cove_{33 \cdot (1, 2)} I_2) \overset{\epsilon \downarrow 0}{\rightarrow} \frac{\epsilon}{2 \pi \cove_{33 \cdot (1, 2)}}.$$
	
	In this representation, the likelihood from~\eqref{eqn:likelihood} factorizes as:
	
	\begin{align*}
	p(\data \given \cov) = &\prod_{i=1}^2 (2\pi)^{-\half{\nobs}} \cove_{ii}^{-\half{\nobs}} \exp\left[-\half{1} \scme_{ii} \cove_{ii}^{-1} \right] \times (2\pi)^{-\half{\nobs}} \cove_{33 \cdot (1, 2)}^{-\half{\nobs}} \exp\left[-\half{1} \scme_{33 \cdot (1,2)} \cove_{33 \cdot (1, 2)}^{-1} \right] \times \\
	& \Gauss(\cov_{(1, 2), (1, 2)}^{-1} \cov_{(1, 2), 3}; \scm_{(1, 2), (1, 2)}^{-1} \scm_{(1, 2), 3}, \cove_{33 \cdot (1,2)} \scm_{(1, 2), (1, 2)}^{-1}) \frac{2\pi \cove_{33 \cdot (1, 2)}}{|\scm_{(1, 2), (1, 2)}|^{\half{1}}}.
	\end{align*}
	
	With some bookkeeping, keeping only leading terms in $\epsilon$, we obtain:
	
	$$ p(\data \given \var_1 \indep \var_2) = \frac{\epsilon^\half{3 \nu}}{\pi^\half{3 \nobs}} \left[ \frac{\Gamma_1(\half{\nnu - 1})}{\Gamma_1(\half{\nu - 1})} \right]^2 \frac{\Gamma_1(\half{\nnu})}{\Gamma_1(\half{\nu})}  \scme_{11}^{-\frac{n + \nu - 1}{2}} \scme_{22}^{-\frac{n + \nu - 1}{2}} \scme_{33\cdot(1, 2)} ^{-\frac{n + \nu}{2}} |\scm_{(1, 2), (1, 2)}|^{-\frac{1}{2}}.$$
	
	Finally, dividing by~\eqref{eqn:ref_evidence}, the Bayes factor in the limit $\epsilon \downarrow 0$ is:
	
	\begin{align*}
		\Bf(\var_1 \indep \var_2) &= \frac{\Gamma_2(\frac{\nu - 1}{2})}{\Gamma_2(\frac{n + \nu - 1}{2})} \left[ \frac{\Gamma_1(\frac{n + \nu - 1}{2})}{\Gamma_1(\frac{\nu - 1}{2})} \right]^2 |\scor_{(1, 2), (1, 2)}|^{\frac{n + \nu - 1}{2}} \\
		&= \frac{\Gamma(\frac{\nnu - 1}{2})}{\Gamma(\frac{\nnu - 2}{2})}  \frac{\Gamma(\frac{\nu - 2}{2})}{\Gamma(\frac{\nu - 1}{2})} (1 - \score_{12}^2)^{\frac{n + \nu - 1}{2}}.
	\end{align*}
	$$ $$

\item We now show how to derive the Bayes factor for the `$\var_1 \indep  \var_2 \given \var_3$' case. The derivation for the other two permutations, `$\var_2 \indep  \var_3 \given \var_1$' and `$\var_3 \indep  \var_1 \given \var_2$', is completely analogous. The inverse-Wishart distribution in~\eqref{eqn:prior_cov} factorizes, for arbitrary $\scl$ and $\nu$, into:
	
	\begin{align*}
	\iWish{3}(\cov; \scl, \nu) = &\iWish{2}(\cov_{(1, 2), (1, 2) \cdot 3}; \scl_{(1, 2), (1, 2) \cdot 3}, \nu) \times \iWish{1}(\cove_{33}; \scle_{33}, \nu - 2) \times \\ & \Gauss(\cove_{33}^{-1} \cov_{3, (1, 2)}; \scle_{33}^{-1} \scl_{3, (1, 2)}, \cove_{(1, 2), (1, 2) \cdot 3} \scle_{33}^{-1}).
	\end{align*}

	We can implement the constraint $\icove_{12} = 0$, where $\icov = \cov^{-1}$ is the precision matrix, by forcing $\cov_{(1, 2), (1, 2) \cdot 3}$ to be diagonal. This suggests the prior:
	
	\begin{align*}
		p(\cov \given \var_1 \indep \var_2 \given \var_3) =& \prod_{i=1}^2 \iWish{1}(\cove_{ii \cdot 3}; \epsilon, \nu) \times \iWish{1}(\cove_{33 }; \epsilon, \nu - 2) \times \\
		& \Gauss(\cove_{33}^{-1} \cov_{3, (1, 2)}; 0, \epsilon^{-1} \cov_{(1, 2), (1, 2) \cdot 3}).
	\end{align*}
	where we use the shorthand notation $\cove_{ii \cdot 3}$ to denote $(\cov_{(1, 2), (1, 2) \cdot 3})_{ii}$ for $i \in \{1, 2\}$ (same for the scatter matrix $\scm$). Note that in the limit $\epsilon \downarrow 0$, the normal component moves towards an improper flat distribution:
	
	$$ \Gauss(\cove_{33}^{-1} \cov_{3, (1, 2)}; 0, \epsilon^{-1} \cov_{(1, 2), (1, 2) \cdot 3}) \overset{\epsilon \downarrow 0}{\rightarrow} \frac{\epsilon}{2 \pi |\cov_{(1, 2), (1, 2) \cdot 3}|^{\half{1}}}.$$
	
	In this representation, the likelihood from~\eqref{eqn:likelihood} factorizes as:
	
	\begin{align*}
		p(\data \given \cov) = &\prod_{i=1}^2 (2\pi)^{-\half{\nobs}} \cove_{ii \cdot 3}^{-\half{\nobs}} \exp\left[-\half{1} \scme_{ii \cdot 3} \cove_{ii \cdot 3}^{-1} \right] \times (2\pi)^{-\half{\nobs}} \cove_{33}^{-\half{\nobs}} \exp\left[-\half{1} \scme_{33} \cove_{33}^{-1} \right] \times \\
		& \Gauss(\cove_{33}^{-1} \cov_{3, (1, 2)}; \scme_{33}^{-1} \scm_{3, (1, 2)}, \cov_{(1, 2), (1, 2) \cdot 3} \scme_{33}^{-1}) \frac{2\pi |\cov_{(1, 2), (1, 2) \cdot 3}|^{\half{1}}}{\scme_{33}}
	\end{align*} 
	
	With some bookkeeping, keeping only leading terms in $\epsilon$, we obtain:
	
	$$ p(\data \given \var_1 \indep \var_2 \given \var_3) = \frac{\epsilon^\half{3 \nu}}{\pi^\half{3 \nobs}} \left[ \frac{\Gamma_1(\half{\nnu})}{\Gamma_1(\half{\nu})} \right]^2 \frac{\Gamma_1(\half{\nnu - 2})}{\Gamma_1(\half{\nu - 2})}  \scme_{11 \cdot 3}^{-\half{\nnu}} \scme_{22 \cdot 3}^{-\half{\nnu}} \scme_{33} ^{-\half{\nnu}}.$$
	
	Finally, dividing by~\eqref{eqn:ref_evidence}, the Bayes factor in the limit $\epsilon \downarrow 0$ is:
	
	\begin{align*}
		\Bf(\var_1 \indep \var_2 \given \var_3) &= \frac{\Gamma_2(\half{\nu})}{\Gamma_2(\half{\nnu})} \left[ \frac{\Gamma_1(\half{\nnu})}{\Gamma_1(\half{\nu})} \right]^2 \left[ \frac{|\scor_{(1, 2), (1, 2) \cdot 3}|}{\score_{11 \cdot 3} \score_{22 \cdot 3}}\right]^{\half{\nnu}} \\
		&= \frac{\Gamma(\half{\nnu})}{\Gamma(\half{\nnu - 1})} \frac{\Gamma(\half{\nu - 1})}{\Gamma(\half{\nu})} \left[\frac{|\scor|}{(1 - \score_{13}^2)(1 - \score_{23}^2)}\right]^{\half{\nnu}}.
	\end{align*}

\end{itemize}

To sum up, we obtain the \textit{Bayes factors on covariance structures} (BFCS):

\begin{equation}
	\begin{aligned} 
	\Bf(\var_1 \indep \var_2 \indep \var_3) &= f(\nobs, \nu) g(\nobs, \nu) |\scor|^{\frac{\nobs + \nu}{2}} \\
	\Bf(\var_3 \indep (\var_1, \var_2)) &= f(\nobs, \nu) \left(\frac{|\scor|}{1 - \score^2_{12}}\right)^\frac{\nobs + \nu}{2} \\
	\Bf(\var_1 \indep \var_2 \given \var_3) &= g(\nobs, \nu) \left( \frac{|\scor|}{(1 - \score_{13}^2)(1 - \score_{23}^2)} \right)^\frac{\nobs + \nu}{2} \\
	\Bf(\var_1 \indep \var_2) &= \frac{f(\nobs, \nu)}{g(\nobs, \nu)} (1 - \score_{12}^2)^\frac{\nobs + \nu - 1}{2},
	\end{aligned} \label{eqn:bayes_factors}
\end{equation}
where $f(\nobs, \nu) = \dfrac{\nobs + \nu - 2}{\nu - 2}$ and $g(\nobs, \nu) = \dfrac{\Gamma\left(\frac{\nobs + \nu}{2}\right) \Gamma\left(\frac{\nu - 1}{2}\right)}{\Gamma\left(\frac{\nobs + \nu - 1}{2}\right) \Gamma\left(\frac{\nu}{2}\right)} \approx \left(\dfrac{2\nobs + 2\nu - 3}{2 \nu - 3}\right)^\frac{1}{2}$.

We see from~\eqref{eqn:bayes_factors} that in order to derive the Bayes factors we only need to plug in the sample correlation matrix $\scor$ with the number of observations $\nobs$ and to compute a limited number of closed-form terms. This is then sufficient to obtain the full posterior distribution over the covariance structures, making the BFCS method fast and efficient.

\subsection{Priors on Causal Structures} \label{ssec:prior_cstr}

To do a full Bayesian analysis, we need to specify priors over the different conditional independence models. Assuming the faithfulness condition~\cite{spirtes_causation_2000}, there is a one-to-one correspondence between the Markov equivalence classes of the underlying causal graph structure and the conditional independence models. By taking a uniform prior over causal graphs and denoting by $|\model_j|$ the number of causal graphs consistent with the independence model $\model_j \in \mathcal{J}$, we arrive at the prior: 

$$p(\model_j) = \frac{|\model_j|}{\sum_i |\model_i|}, 
\quad \forall \model_j \in \mathcal{J}.$$

\begin{table}[!htb]
	\centering \scriptsize
	\begin{tabular}{ccc|cccc}
		Case & CI Model & Description & DAG & DAG w/ BK & DMAG & DMAG w/ BK \\
		\hline \hline
		Full & $\model_0$ & $\var_1 \nindep \var_2 \nindep \var_3$ & 6 & 2 & 19 & 3 \\ \hline
		& $\model_1$ & $\var_1 \indep \var_2$ & 1 & 1 & 3 & 2 \\
		Acausal & $\model_2$ & $\var_2 \indep \var_3$ & 1 & 0 & 3 & 0 \\
		& $\model_3$ & $\var_3 \indep \var_1$ & 1 & 1 & 3 & 2 \\ \hline
		& $\model_4$ & $\var_1 \indep \var_2 \given \var_3$ & 3 & 1 & 5 & 1 \\
		Causal & $\model_5$ & $\var_2 \indep \var_3 \given \var_1$ & 3 & 1 & 5 & 1 \\
		& $\model_6$ & $\var_3 \indep \var_1 \given \var_2$ & 3 & 1 & 5 & 1 \\ \hline
		& $\model_7$ & $\var_1 \indep (\var_2, \var_3)$ & 2 & 2 & 3 & 3 \\
		Independent& $\model_8$ & $\var_2 \indep (\var_3, \var_1)$ & 2 & 1 & 3 & 1 \\
		& $\model_9$ & $\var_3 \indep (\var_1, \var_2)$ & 2 & 1 & 3 & 1 \\ \hline
		Empty & $\model_{10}$ & $\var_1 \indep \var_2 \indep \var_3$ & 1 & 1 & 1 & 1 \\ \hline \hline
		All & &  & 25 & 12 & 53 & 16 \\
	\end{tabular}
	\caption{Number of causal graph structures over three variables for each conditional independence model (equivalently, in each Markov equivalence class). In the columns marked `w /BK', the background knowledge that $\var_1$ precedes all other variables, i.e., there can be no arrowhead towards $\var_1$, is added when counting the number of structures.}
	\label{tab:no_str_3var}
\end{table}

In Table~\ref{tab:no_str_3var} we count the number of DAGs and DMAGs (see Section~\ref{sec:bkg}) consistent with each conditional independence model. The addition of background knowledge (BK) reduces the number of causal graph structures corresponding to each conditional independence model. Specifically relevant for discovering causal regulatory relationships is the background knowledge that the genetic marker precedes the expression traits, i.e., that $X_1$ precedes all other variables. This additional constraint leads to the counts in the columns marked `w/ BK' in Table~\ref{tab:no_str_3var}. Some conditional independence models, or equivalently some covariance structures, imply acausal or causal statements (Figure~\ref{fig:equivalence}), which is what allows us to directly translate the posterior probabilities over covariance structures into statements over causal relationships.

In this paper, we have proposed simple uniform priors on causal structures from which we derive priors on conditional independence models. For example, if we assume an underlying DAG structure without any background knowledge, then $p(\var_3 \indep \var_1 \given \var_2) = \frac{3}{25}$ according to the counts in Table~\ref{tab:no_str_3var}. The approach also allows for the addition of more specific background knowledge. For example, if we know that a certain causal regulatory relationship $\var_i \to \var_j$ cannot occur, we can set the probability of all causal graphs containing that directed edge to zero. Conversely, we can incorporate a previously established causal regulatory relationship by setting the prior probability of causal graphs not containing that particular relationship to zero. This prior knowledge can come from literature or databases such as KEGG~\cite{kanehisa_kegg_2000}. Other types of background knowledge, such as information regarding the sparsity of a GRN, can also be incorporated. One idea is to introduce a parameter $q$ indicating the probability of a (direct) causal regulatory relationship between two genes. If we then assume that each edge occurs independently, without taking direction into consideration, we simply compute the product of the edge probabilities for each causal graph: the prior probability of the `full graph' would then be proportional to $q^3$, while the prior probability of the `empty graph' would be proportional to $(1-q)^3$.

Now that we have defined priors on the conditional independence models (Markov equivalence classes), we can derive the posterior probabilities from equations~\eqref{eqn:post_ci_model} and~\eqref{eqn:bayes_factors}. We are mainly interested in finding the causal structure $\var_1 \to \var_2 \to \var_3$, which corresponds to the LCD pattern~\cite{cooper_simple_1997}. Assuming DAGs or DMAGs with the background knowledge that $\var_1$ precedes all other variables, there is a one-to-one correspondence between the causal structure and the conditional independence model $\model_6$, so we want to estimate $p(\model_6 \given \data)$.

For each prior, we can compute a different upper bound on the probability $p(\model_6 \given \data)$, which represents a limit on the confidence in this statement given a particular number of samples. From Equation~\eqref{eqn:post_ci_model}, we derive the  upper bound

$$ p(\model_6 \given \data) = \frac{\Bf(\model_6) \cdot p(\model_6)}{\sum_{i=0}^{10} \Bf(\model_i) \cdot p(\model_i)} \le \frac{\Bf(\model_6) \cdot p(\model_6)}{\Bf(\model_6) \cdot p(\model_6) + p(\model_0)}.$$ Furthermore, we have that for any correlation matrix $\scor$:

$$ |\scor| = 1 + 2 \score_{12} \score_{13} \score_{23} - \score_{12}^2 - \score_{13}^2 - \score_{23}^2 \le (1 - \score_{12}^2)(1 - \score_{13}^2).$$ It is easy to show that this inequality is equivalent to $(\score_{12} \score_{23} - \score_{13})^2 \ge 0$. Combining this result with the Bayes factor expression for $\model_6$ (`$\var_1 \indep \var_3 \given \var_2$') in~\eqref{eqn:bayes_factors}, we get $\Bf(\model_6) \le g(\nobs, \nu)$, which in turn implies

\begin{equation}
p(\model_6 \given \data) \le \frac{\Bf(\model_6) \cdot p(\model_6)}{\Bf(\model_6) \cdot p(\model_6) + p(\model_0)} \le \frac{g(\nobs, \nu) \cdot p(\model_6)}{g(\nobs, \nu) \cdot p(\model_6) + p(\model_0)},
\end{equation}
where $p(\model_6)$ is the prior probability of the conditional independence model `$\var_1 \indep \var_3 \given \var_2$', $p(\model_0)$ is the prior probability of the reference model `$\var_1 \nindep \var_2 \nindep \var_3$' and $g(\nobs, \nu) = \dfrac{\Gamma\left(\frac{\nobs + \nu}{2}\right) \Gamma\left(\frac{\nu - 1}{2}\right)}{\Gamma\left(\frac{\nobs + \nu - 1}{2}\right) \Gamma\left(\frac{\nu}{2}\right)} \approx \left(\dfrac{2\nobs + 2\nu - 3}{2 \nu - 3}\right)^\frac{1}{2}$. This result means that the posterior probability of `$\var_1 \indep \var_3 \given \var_2$' is upper bounded by a quantity that depends on the number of observations and on the prior belief in `$\var_1 \indep \var_3 \given \var_2$' versus the full model.

The posterior probabilities could also be derived by combining the Bayesian Gaussian equivalent (BGe) score~\citep{geiger_learning_1994} with the priors on causal structures defined in this subsection. Due to our choice of priors on covariance matrices (see Subsection~\ref{ssec:choosing_prior_covm}), however, the Bayes factors are simpler to compute. This makes our approach more efficient when used to infer causal relationships in large regulatory networks. 

To summarize, we have developed a method for computing the posterior probabilities of the covariance structures over three variables (BFCS). We will employ this procedure as part of an algorithm for discovering regulatory relationships. The idea is to search over triplets of variables to find potential local causal structures, in a similar way to the LCD and Trigger algorithms (see Algorithm~\ref{alg:BFCS_yeast}).

\subsection{Finding Causal Links Using BFCS} \label{ssec:finding_causal_links}

Now that we have derived the posterior probability of the causal structure p($\var_1 \to \var_2 \to \var_3 \given \data$), we would like to estimate the probability of the causal link $\var_2 \to \var_3$. There are multiple ways of arriving at an estimate of this link. In Trigger, this is handled by limiting the search to the variable $\var_1$ that has the (locally) strongest primary linkage to $\var_2$ and then using the estimate for $p(\var_1 \to \var_2 \to \var_3 \given \data)$ as the posterior probability of the causal link. This selection strategy will reduce the search space significantly, but it does not necessarily lead to the best estimate, because the variable with the strongest primary linkage might also be directly linked to $\var_3$. Instead, we propose to derive the posterior probability of $\var_k \to \var_2 \to \var_3$ given the data for each available $X_k$ with our much more efficient approach and take the maximum of these probabilities over $k$. From the Fr\'echet inequalities, we see that this constitutes a lower bound on the probability $p(\lor_k \var_k \to \var_2 \to \var_3 \given \data)$~\cite{sokolova_computing_2016}, which is the posterior probability of finding the local structure $\var_k \to \var_2 \to \var_3$ for at least one $k$, which in turn is a lower bound on $p(\var_2 \to \var_3 \given \data)$. Consequently, in a similar vein to the authors of Trigger, we report $\max_k p(\var_k \to \var_2 \to \var_3 \given \data)$ as a conservative estimate for the probability of $\var_2 \rightarrow \var_3 \given \data$. This means we not only have a different way of estimating the posterior probabilities of causal structures, but we also employ a different search strategy when translating these into the probability of a causal link.

\section{Experimental Results} \label{sec:experiments}

\subsection{Consistency of Detecting Local Causal Structures}

\begin{figure}[!htb]
	\begin{subfigure}{.33\linewidth}
		\includegraphics{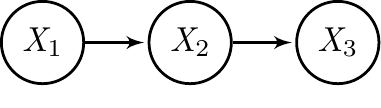}
		\subcaption{Causal model} \label{subfig:generating_causal}
	\end{subfigure} %
	\begin{subfigure}{.33\linewidth}
		\includegraphics{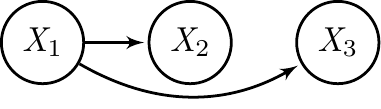}
		\subcaption{Independent model}
	\end{subfigure} %
	\begin{subfigure}{.32\linewidth}
		\includegraphics{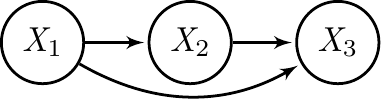}
		\subcaption{Full model}
	\end{subfigure}
	\caption{Generating models} \label{fig:generating_models}
\end{figure}

In this simulation, we assessed how well our BFCS approach is able to detect the causal structure $\var_1 \rightarrow \var_2 \rightarrow \var_3$, which is crucial to the application of the LCD and Trigger algorithms. We considered the three generating structures depicted in Figure~\ref{fig:generating_models}. In all three cases, the variables are mutually marginally dependent, but only in the first model $\var_1 \indep \var_3 \given \var_2$ holds. Note that the labels given to the generating models in Figure~\ref{fig:generating_models} are commonly used in the genetics and genomics literature (see, e.g., Figure 1 in~\cite{neto_modeling_2013}), but are different and should not be confused with the labels we have assigned to the canonical cases from Figure~\ref{fig:equivalence} and Table~\ref{tab:no_str_3var}. 

We sampled data from a linear structural equation model over $(\var_1, \var_2, \var_3)$:
\begin{equation}
	\label{eqn:sem_consistency}
	\begin{aligned}
		\var_1 &:= \varepsilon_1; \\
		\var_2 &:= \varepsilon_2 + b_{21} \var_1; \\
		\var_3 &:= \varepsilon_3 + b_{31} \var_1 + b_{32} \var_2.
	\end{aligned}
\end{equation}
The linear structural parameters (interaction strengths), denoted by $b$, express the strength of the causal relationships and are set to zero in the absence of a causal link: $b_{31} := 0$ in the `causal model' and $b_{32} := 0$ in the `independent model' from Figure~\ref{fig:generating_models}. The nonzero structural parameters were sampled from independent standard normal distributions.

Given each generating model in Figure~\ref{fig:generating_models}, we generated data sets of different sizes (from $10^2$ up to $10^6$ samples). For each data set, we computed the correlation matrix, which we plugged into~\ref{eqn:bayes_factors} for computing the Bayes factors. We repeated this procedure for 1000 different parameter configurations. We assumed that $\var_1$ precedes all other variables and we allowed for latent variables, i.e., we did not use the knowledge that the data is causally sufficient. We considered a uniform prior over the sixteen possible DMAG structures (see Table~\ref{tab:no_str_3var}). 

\begin{figure}[!htb]
	\begin{subfigure}{.33\linewidth}
		\includegraphics[width = \textwidth]{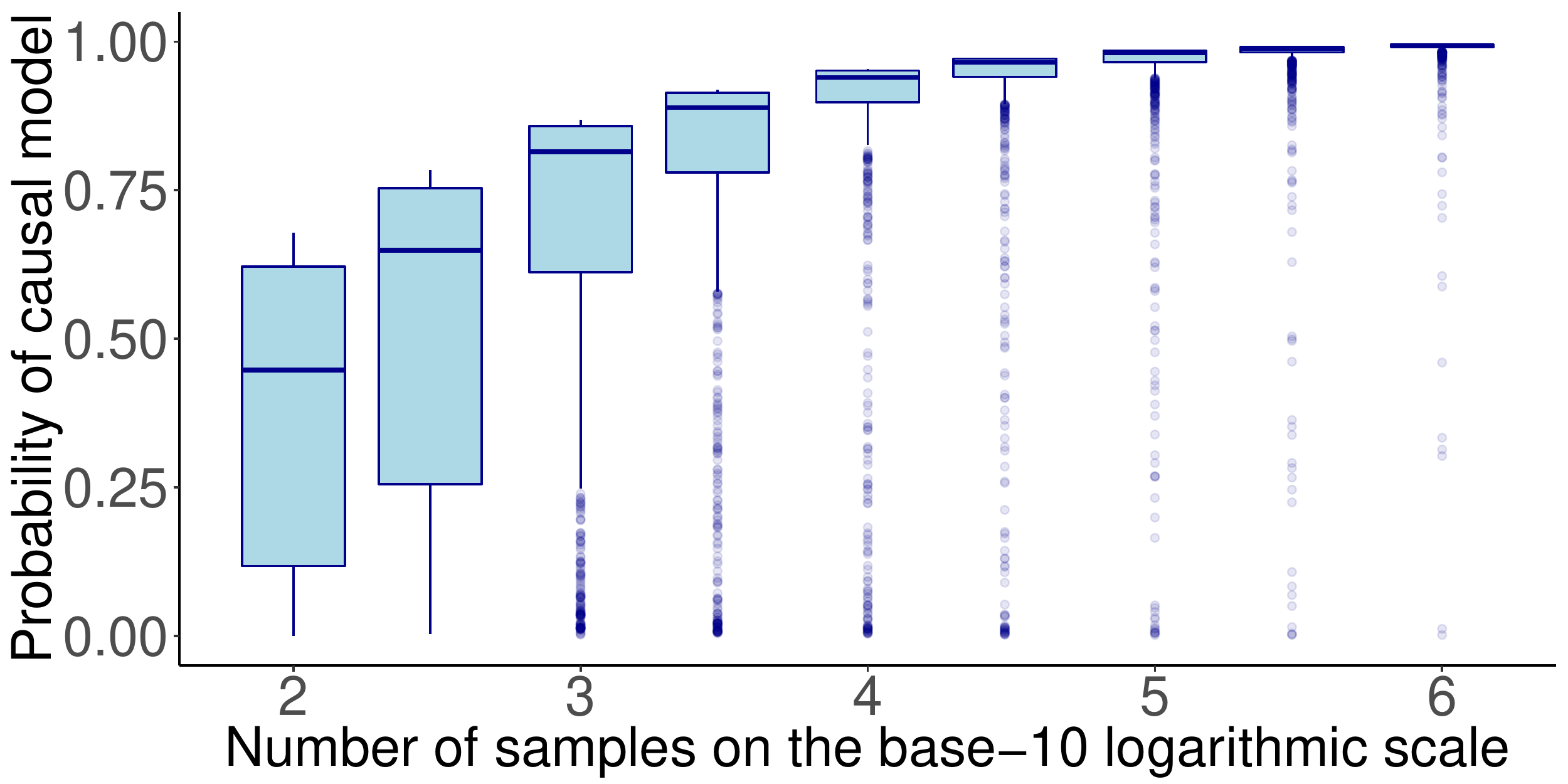}
		\subcaption{Causal model}
		\label{subfig:consistency_T_norm}
	\end{subfigure}
	\begin{subfigure}{.33\linewidth}
		\includegraphics[width = \textwidth]{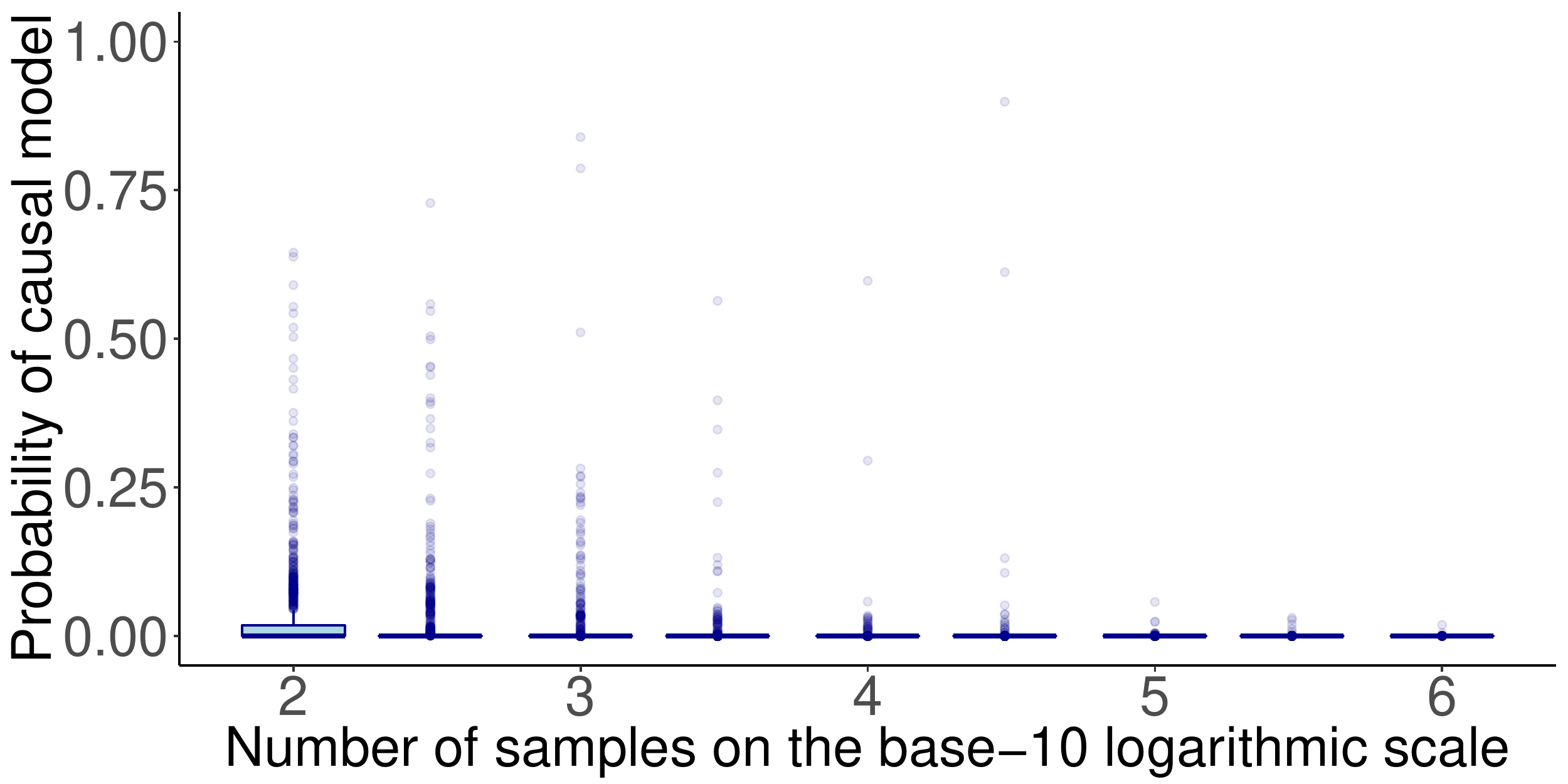}
		\subcaption{Independent model}
	\end{subfigure}
	\begin{subfigure}{.32\linewidth}
		\includegraphics[width = \textwidth]{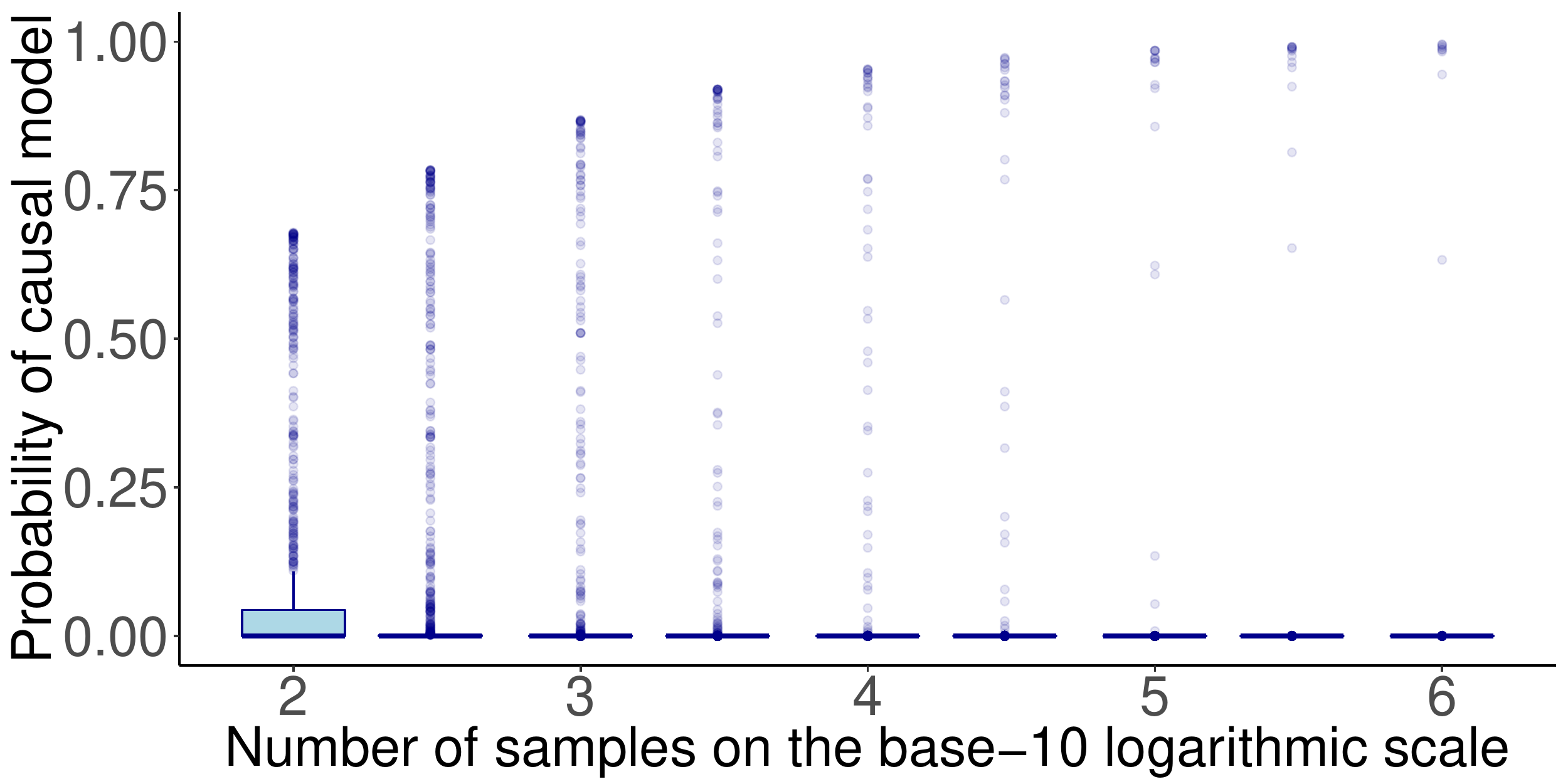}
		\subcaption{Full model}
	\end{subfigure}
	\begin{subfigure}{.33\linewidth}
		\includegraphics[width = \textwidth]{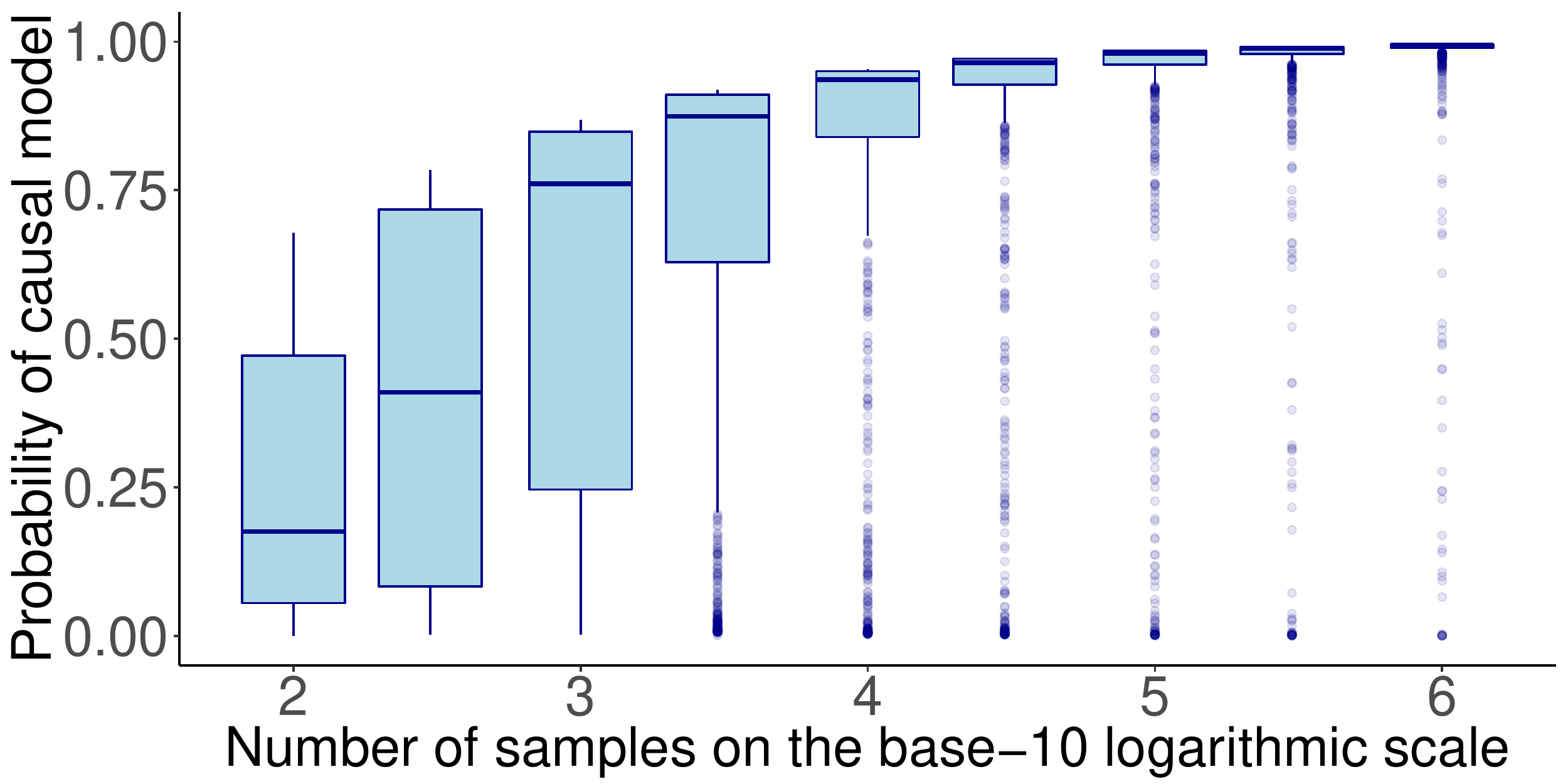}
		\subcaption{Causal model} \label{subfig:consistency_T_binom}
	\end{subfigure}
	\begin{subfigure}{.33\linewidth}
		\includegraphics[width = \textwidth]{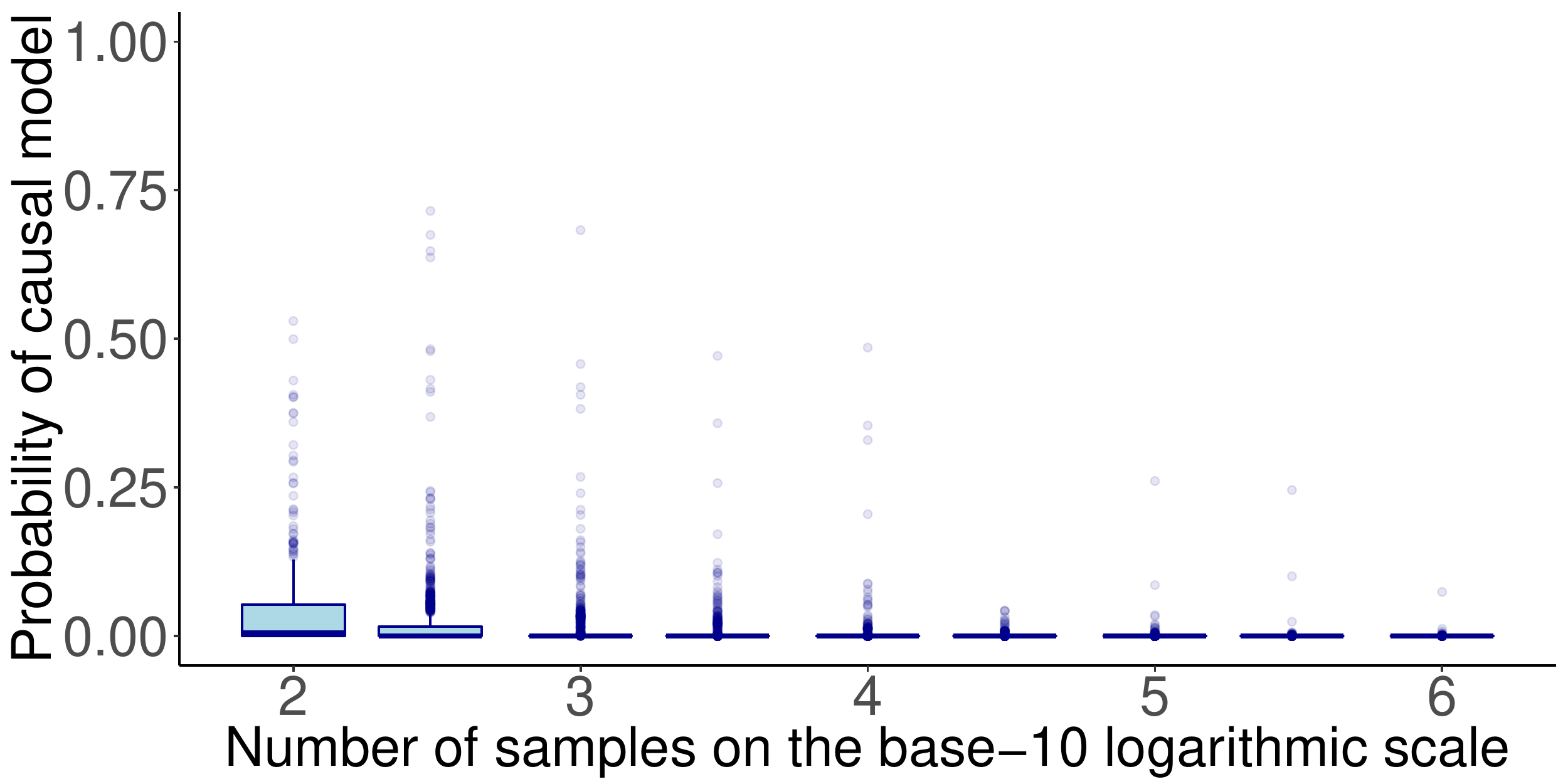}
		\subcaption{Independent model}
	\end{subfigure}
	\begin{subfigure}{.32\linewidth}
		\includegraphics[width = \textwidth]{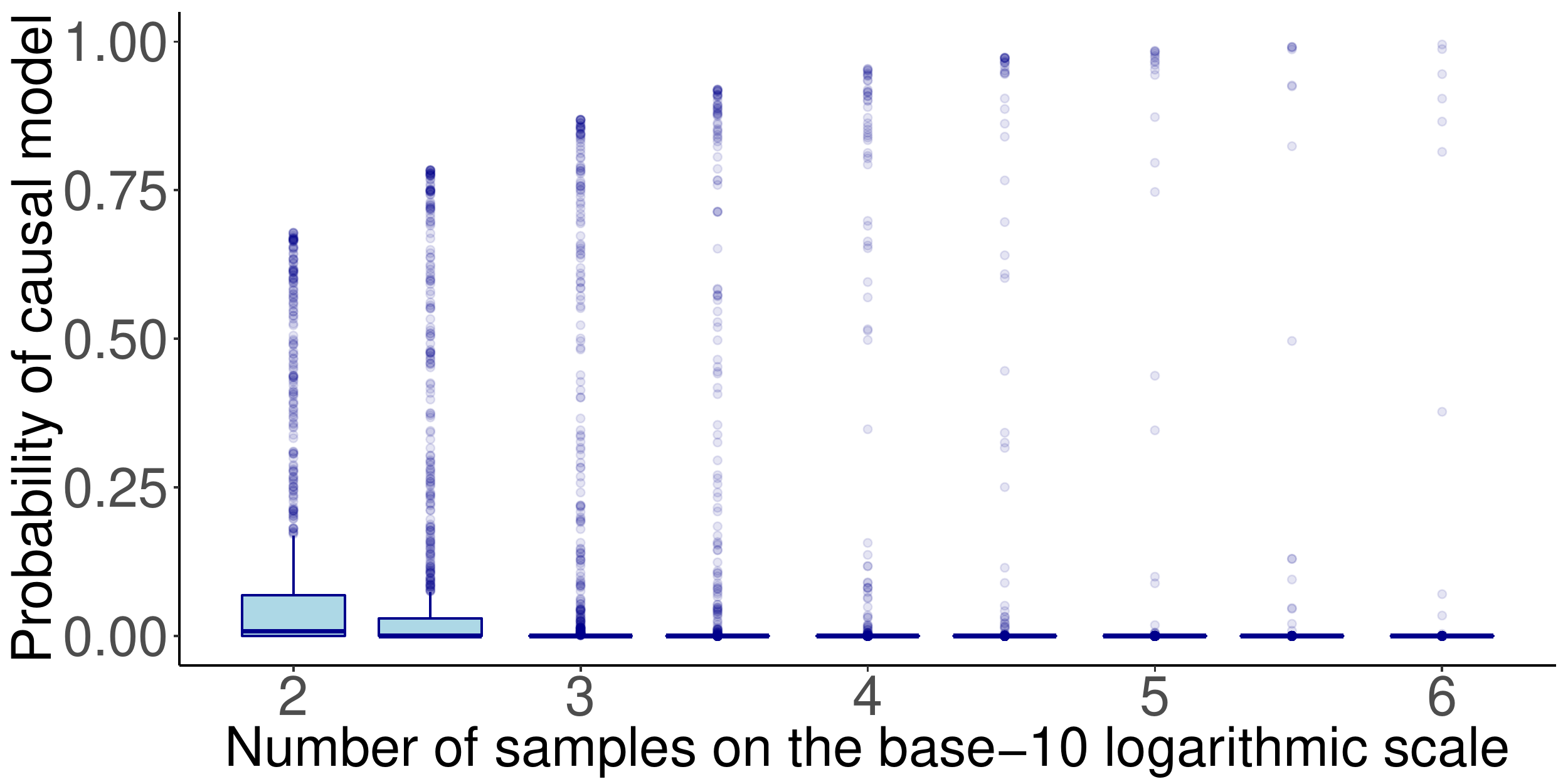}
		\subcaption{Full model}
	\end{subfigure}
	\caption{Box plots of the posterior probabilities of $p(\var_1 \rightarrow \var_2 \rightarrow \var_3 \given \data)$ output by BFCS across the 1000 different parameter configurations for each of the generating models in Figure~\ref{fig:generating_models}. As we increased the number of samples, the probability $p(\var_1 \rightarrow \var_2 \rightarrow \var_3 \given \data)$ converged to one when the data was generated from the causal model (a) and converged to zero when it was not (b and c). \textbf{Top}: $\varv = (\var_1, \var_2, \var_3)$ is multivariate Gaussian; \textbf{Bottom}: $\var_1$ is Bernoulli and $(\var_2, \var_3) \given \var_1$ is Gaussian.} \label{fig:consistency}
\end{figure}

In the first experiment we generated random multivariate data for each of the models in Figure~\ref{fig:generating_models} by sampling independent standard Gaussian noise, that is $\varepsilon_1, \varepsilon_2, \varepsilon_3 \sim \mathcal{N}(0, 1)$ and then deriving $\var_1, \var_2, \var_3$ from the structural equation model in~\eqref{eqn:sem_consistency}. As expected, the posterior probability $p(\var_1 \rightarrow \var_2 \rightarrow \var_3 \given \data)$ estimated with BFCS converged to one (Figure~\ref{fig:consistency}, top row) when the true generating model was the one in Figure~\ref{subfig:generating_causal}. At the same time, the estimate for $p(\var_1 \rightarrow \var_2 \rightarrow \var_3 \given \data)$ converged to zero when the true generating model was the independent or full model. 

Note that in our simulation it is easier to distinguish the causal model from the independent model than from the full model. When generating data from the full model, it is possible to generate structural parameters that are close to zero. If the direct interaction between $\var_1$ and $\var_3$ is close to zero ($b_{31} \approxeq 0$), it looks as if $\var_1 \indep \var_3 \given \var_2$. For example, in one of the repetitions we sampled the structural parameters $b_{21} = -0.103, b_{32} = 1.467, b_{31} = -4.175 \cdot 10^{-4}$, in which case the influence of $\var_1$ on $\var_3$ is too small to be detected, even from a data set of $10^6$ samples, effectively reducing the `full model' to the `causal model'. However, the direct interaction from $\var_1$ to $\var_3$ will eventually be detected given enough data samples.

In the second experiment, we considered the same generating models from Figure~\ref{fig:generating_models}, but we instead sampled $\varepsilon_1$ $(\var_1)$ from a Bernoulli distribution to mimic a genetic variable, e.g., an allele or the parental strain in the yeast experiment~\cite{chen_harnessing_2007}. We sampled a different success probability $p$ for the Bernoulli variable $\varepsilon_1 \sim B(1, p)$ in each repetition from a uniform distribution between 0.1 and 0.5. The other two noise terms were sampled from independent standard Gaussian distributions like before: $\varepsilon_2, \varepsilon_3 \sim \mathcal{N}(0, 1)$. We then derived $\var_1, \var_2, \var_3$ from the structural equation model in~\eqref{eqn:sem_consistency}. In the bottom row of Figure~\ref{fig:consistency} we see that when the Gaussian assumption did not hold for $\var_1$, we lost some power in recovering the correct model. When we increased the number of samples, this loss in power due to the violation of the Gaussian assumption became less severe and BFCS remained consistent.

\subsection{Causal Discovery in Gene Regulatory Networks}

In this series of experiments, we simulated transcriptional regulatory networks meant to emulate the yeast data set analyzed in~\cite{chen_harnessing_2007}. We randomly generated data for 100 genetic markers, where each marker is an independent Bernoulli variable with `success' probability uniformly sampled between 0.1 and 0.5. We then generated transcript level expression data from the structural equation model (SEM):

\begin{equation} \label{eqn:grn_sem}
	\textbf{t} := \textbf{B} \textbf{t} + \textbf{A} \textbf{l} + \bm{\varepsilon},
\end{equation}
where $\textbf{t} = \left(T_1, T_2, ..., T_{100}\right)^\trp$ is the random vector of the expression trait data, $\textbf{l} = \left(L_1, L_2, ..., L_{100}\right)^\trp$ is the random vector of the genetic markers, $\textbf{B}$ is a lower triangular matrix describing the regulatory network structure, $\textbf{A}$ is the matrix describing the causal links between the genetic markers and the expression traits, and $\bm{\varepsilon} \sim \mathcal{N}(\bfzero_{100}, \idm_{100})$ is added noise.

The true causal relationships between gene expression levels are encoded in the directed graph structure defined by $\textbf{B}$, which has been randomly generated using the \texttt{randomDAG} function from the R \texttt{pcalg} package~\cite{kalisch_causal_2012}. In the yeast data set each genetic marker is on average linked to 5.79 expression traits, where we used an absolute correlation coefficient threshold of 0.5 to define `linked'. To arrive at a similar number of links per genetic marker, we randomly sampled the nonzero terms of $\textbf{A}$ from a binomial distribution with 0.05 success probability, which means that each genetic marker in the simulated GRN is on average connected to $100 \times 0.05 = 5$ expression traits. The strengths of the nonzero terms in $\textbf{A}$ and $\textbf{B}$ were sampled from a $\unif(-1, 1)$ distribution. We generated data from two simulated gene regulatory networks on 100 expression traits, given 100 genetic markers: one sparser (54 directed edges) and one denser (247 directed edges).

We can see the causal discovery problem as a binary classification task, where the positive class (the event) is the presence of a directed edge (causal regulatory relationship) from gene $i$ to gene $j$ ($T_i \to T_j$). In Equation~\eqref{eqn:grn_sem}, this corresponds to a nonzero element in the matrix $\mathbf{B}$, i.e., $T_i \to T_j \iff B_{ji} \ne 0$, for any pair of nodes $(i, j) \in \{1, 2, ..., 100\}^2$, $i \ne j$. BFCS, as well as comparable algorithms, output predicted scores for each causal regulatory relationship that represent probability values for the presence of a directed edge. We evaluated the performance of BFCS and compared it to other approaches in terms of the Receiving Operating Characteristic (ROC), Precision-Recall (PRC), and calibration curves. The last curve describes how consistent model probabilities are with the observed event rates. To construct the calibration plot, we sort the predicted probabilities and distribute them in (five) bins. The average predicted probability in each bin is called the `Bin Midpoint Average Estimated Percentage'. For each bin, we then evaluate the `Observed Event Percentage', which is the fraction of events (positives). Ideally, the two percentages are equal for each bin, meaning that a well-calibrated algorithm exhibits a calibration curve close to the diagonal axis.

We included three versions of the BFCS approach in the comparison. In the first two versions, we take the maximum over all probability estimates of $\hat{p}(L_k \to T_i \to T_j \given \data)$ as our conservative estimate for $p(T_i \to T_j \given \data)$. The only difference between these two versions is in the prior on causal structures: a uniform prior on DAGs (`BFCS DAG') and DMAGs (`BFCS DMAG'), respectively, both incorporating the background knowledge that $L_k$ cannot be caused by $T_i$ or $T_j$. In the third version of BFCS (`BFCS loclink'), we first used the selection strategy of Trigger (implemented as \texttt{trigger.loclink} in~\cite{chen_trigger_2017}) to choose a candidate genetic marker $L_i$ exhibiting the strongest primary linkage to $T_i$ among all available genetic markers. We then used BFCS to estimate $\hat{p}(L_i \to T_i \to T_j \given \data)$ and reported this quantity as our conservative estimate for $p(T_i \to T_j \given \data)$. As a reference, we included an equivalent method based on the `Bayesian Gaussian equivalent' (BGe) score in the comparison, for which we used the same prior on causal structures as `BFCS DMAG'. To ensure a fairer comparison, we scaled and centered the data sets generated from the GRNs before computing the BGe score, since BFCS does not make use of any information regarding the location or scale of the data.

\begin{figure}[p]
	\centering 	\vspace{-3cm}
	\includegraphics[width=\linewidth]{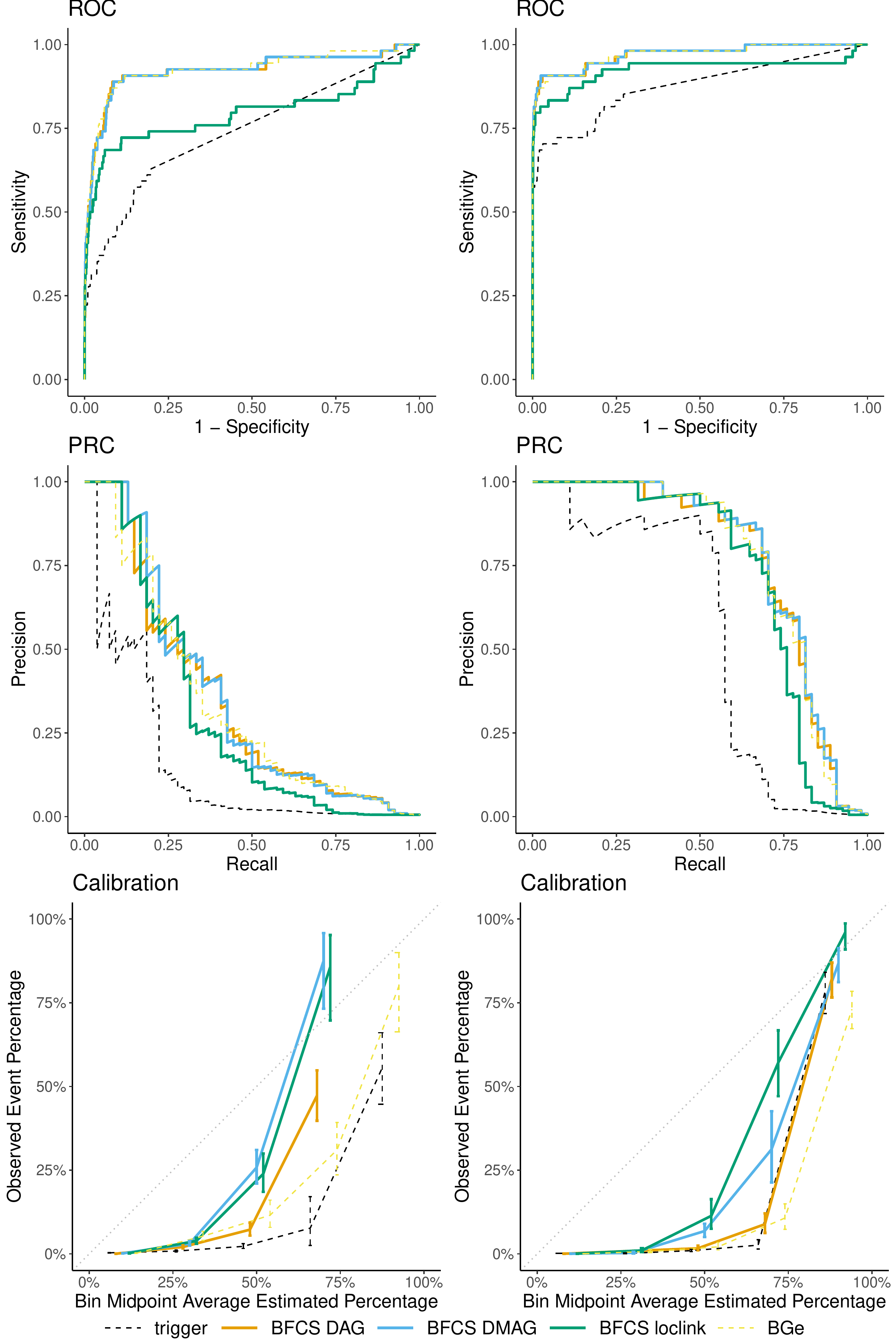}
	\caption{Evaluating the performance in detecting the 54 direct causal regulatory relationships of a (sparser) simulated GRN in terms of ROC (\textbf{top row}), PRC (\textbf{middle row}), and calibration (\textbf{bottom row}) from 100 samples (\textbf{left column}) and 1000 samples (\textbf{right column}). We ran Trigger three times on the simulated data and averaged the results (`trigger') to account for differences when sampling the null statistics. We report the results of three BFCS versions: two described in Algorithm~\ref{alg:BFCS_yeast} for which we take a uniform prior over DAGs (`BFCS DAG') and DMAGs (`BFCS DMAG'), respectively, and one (`BFCS loclink') in which we use the Trigger local-linkage selection strategy described in Subsection~\ref{ssec:finding_causal_links}. For reference, we also show the performance of an equivalent method that uses the BGe score (`BGe').}
	\label{fig:plot_sG_dir}
\end{figure}

\begin{figure}[p]
	\centering \vspace{-3cm}
	\includegraphics[width=\linewidth]{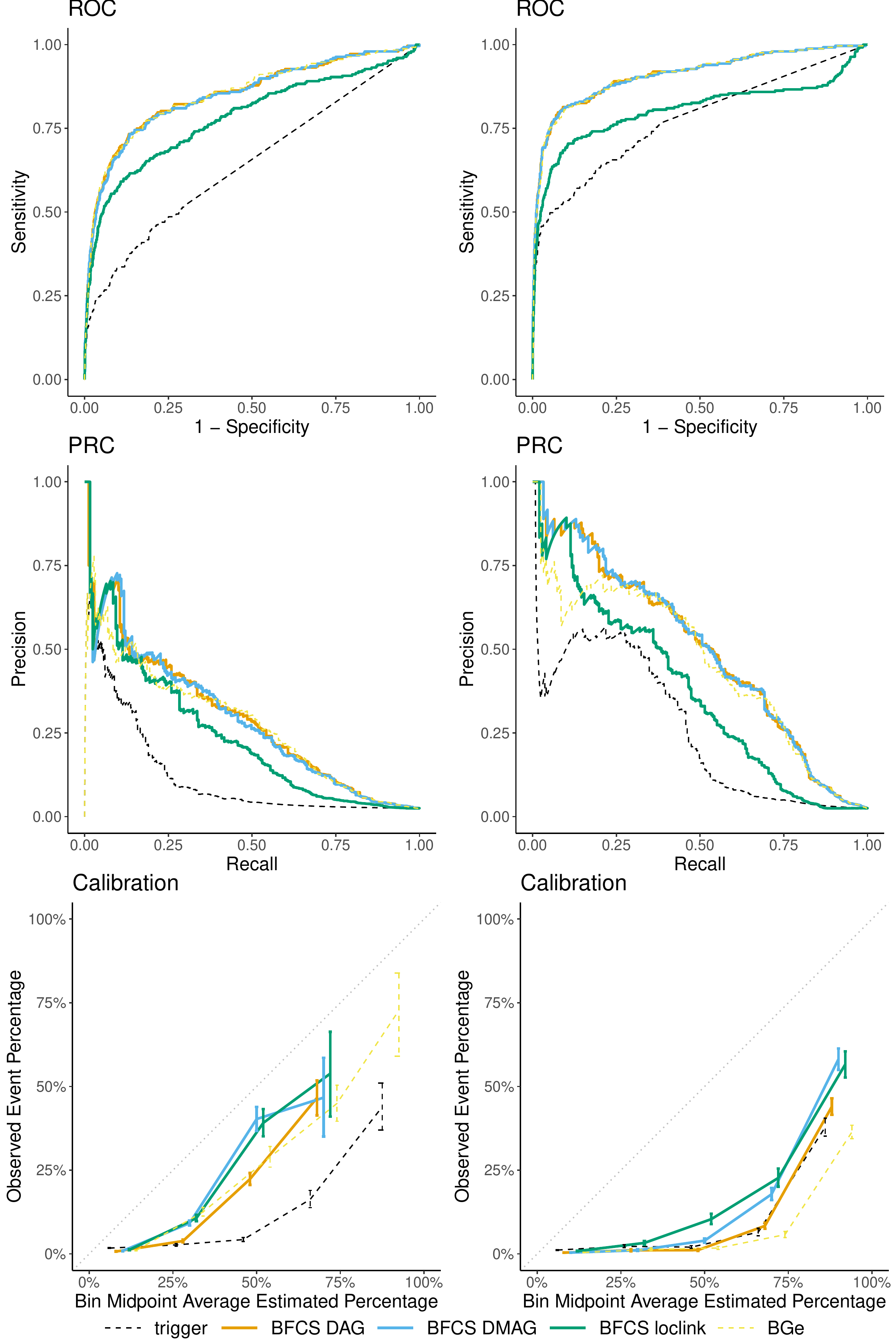}
	\caption{Evaluating the performance in detecting the 247 direct causal regulatory relationships of a (denser) simulated GRN in terms of ROC (\textbf{top row}), PRC (\textbf{middle row}), and calibration (\textbf{bottom row}) from 100 samples (\textbf{left column}) and 1000 samples (\textbf{right column}). We ran Trigger three times on the simulated data and averaged the results (`trigger') to account for differences when sampling the null statistics. We report the results of three BFCS versions: two described in Algorithm~\ref{alg:BFCS_yeast} for which we take a uniform prior over DAGs (`BFCS DAG') and DMAGs (`BFCS DMAG'), respectively, and one (`BFCS loclink') in which we use the Trigger local-linkage selection strategy described in Subsection~\ref{ssec:finding_causal_links}. For reference, we also show the performance of an equivalent method that uses the BGe score (`BGe').} 
	\label{fig:plot_lG_dir}
\end{figure}

We first evaluated how well the algorithms under comparison perform in detecting direct causal regulatory relationships, i.e., those corresponding precisely to the directed edges in the generating network structure (\Cref{fig:plot_sG_dir,fig:plot_lG_dir}). All methods performed better when the underlying network structure was sparser, since there are fewer connections in the graph that can lead to conditional independences resulting from path cancellations. When comparing BFCS to Trigger, we see that all BFCS versions show an improvement in the AUC measure for the ROC and precision-recall curves, in particular `BFCS DAG' and `BFCS DMAG'. We obtained a significantly higher precision for the first causal relationships that are recalled, especially when the data sample size was lower. This suggests that BFCS is better at ranking the top causal regulatory relationships (see also Subsection ~\ref{ssec:yeast}). All BFCS versions, in particular `BFCS loclink' are also better calibrated than Trigger. However, it is important to note that both the BFCS and Trigger approaches are calibrated on detecting the local causal structure $L_k \to T_i \to T_j$, not on finding the causal link $T_i \to T_j$. Trigger in particular lies well under the diagonal line, indicating that its estimated probabilities are consistently overly optimistic. The BGe reference method performs similarly to BFCS in terms of ROC and PRC, but is less well calibrated.

We notice that `BFCS loclink', which employs the same genetic marker search strategy as Trigger for deriving the probability of a causal link given the data, performs better than the other two BFCS versions in terms of calibration, but worse in terms of ROC and PRC. When computing the maximum over posterior probability estimates for many local causal structures, like in `BFCS DAG' and `BFCS DMAG', we expect that the calibration will deteriorate. The more causal structures we look at, the higher the chance that we will accidentally find a triple where the conditional independence we are searching for approximately holds. However, it is still unlikely that BFCS will return a very high probability estimate for this `accidental' conditional independence, which is why the calibration is mainly affected in the bins where these estimates are smaller. At the same time, by considering all triples, the chance of finding a relevant genetic marker for which $L_k \to T_i \to T_j$ is the true local causal structure increases for BFCS. The estimates for these triples will be relatively high, so `BFCS DAG' and `BFCS DMAG' will exhibit a boost in precision for the causal links with the highest estimated posterior probabilities.

Even though we may be mostly interested in direct causal regulatory relationships, the BFCS and Trigger approaches are designed to look for a conditional independence $L_k \indep T_j \given T_i$, which may mean a direct link from $T_i$ to $T_j$ or a causal (regulatory) chain comprised of more than two variables (expression traits). Nevertheless, it is possible to perform a subsequent mediation analysis in order to distinguish between direct and indirect causal links. We can first scan the genome for potential mediator genes $m$ of a discovered causal relationship from gene $i$ to gene $j$. This can be accomplished by selecting those genes for which BFCS reports high probability estimates for both $p(T_i \to T_m \given \data)$ and $p(T_m \to T_j \given \data)$, in addition to a high probability estimate for $p(T_i \to T_j \given \data)$. We then simply check for conditional independences of the form $T_i \indep T_j \given T_m$. If we find this conditional independence, it means that the regulatory relationship from $i$ to $j$ is mediated by $m$. Otherwise, we cannot rule out that there is a direct component in the relationship. The same analysis can be performed for sets of mediators instead of a single mediator.

\begin{figure}[p]
	\centering \vspace{-3cm}
	\includegraphics[width=\linewidth]{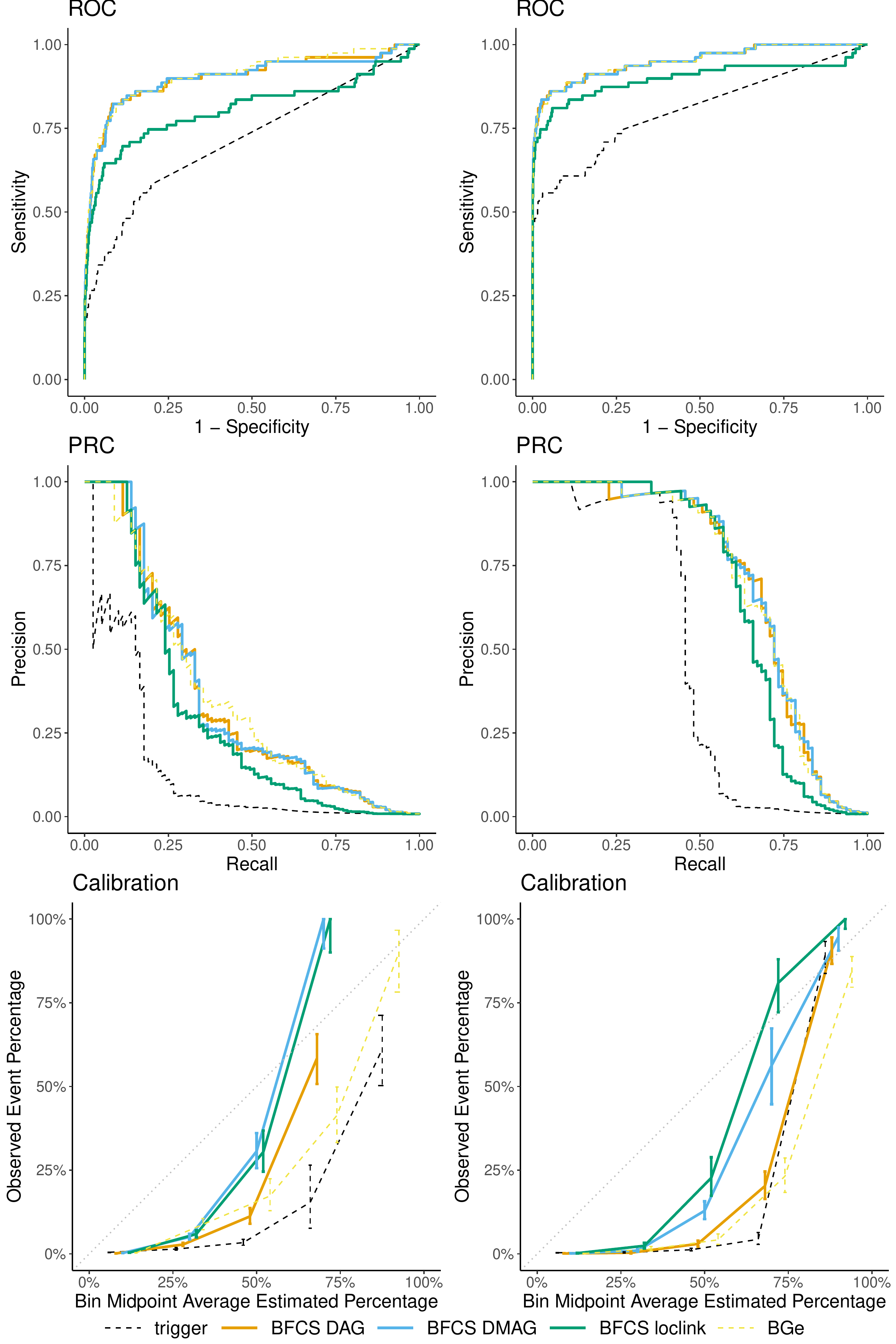}
	\caption{Evaluating the performance in detecting the 79 ancestral causal regulatory relationships of a (sparser) simulated GRN in terms of ROC (\textbf{top row}), PRC (\textbf{middle row}), and calibration (\textbf{bottom row}) from 100 samples (\textbf{left column}) and 1000 samples (\textbf{right column}). We ran Trigger three times on the simulated data and averaged the results (`trigger') to account for differences when sampling the null statistics. We report the results of three BFCS versions: two described in Algorithm~\ref{alg:BFCS_yeast} for which we take a uniform prior over DAGs (`BFCS DAG') and DMAGs (`BFCS DMAG'), respectively, and one (`BFCS loclink') in which we use the Trigger local-linkage selection strategy described in Subsection~\ref{ssec:finding_causal_links}. For reference, we also show the performance of an equivalent method that uses the BGe score (`BGe').} \label{fig:plot_sG_anc}
\end{figure}

\begin{figure}[p]
	\centering \vspace{-3cm}
	\includegraphics[width=\linewidth]{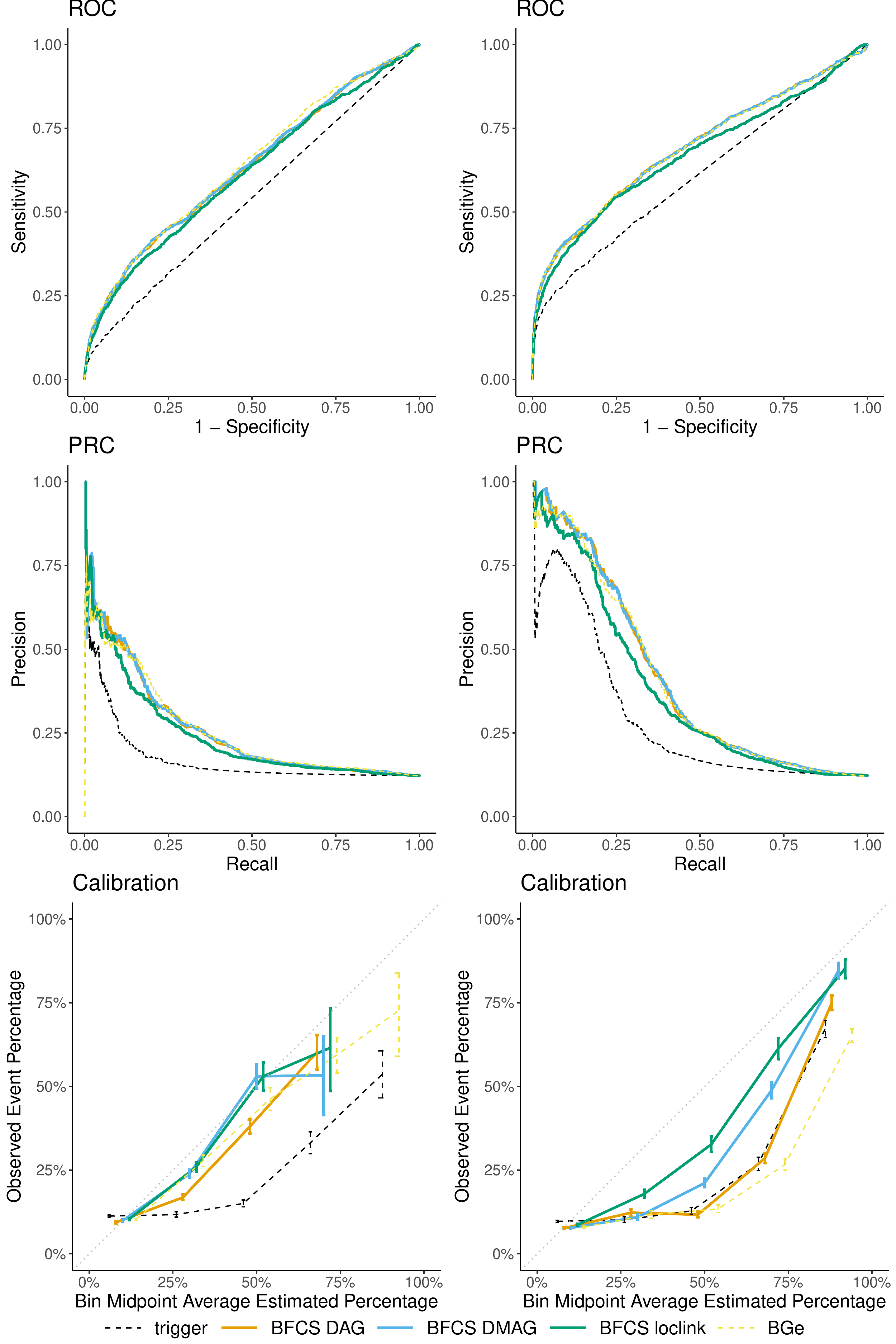}
	\caption{Evaluating the performance in detecting the 1211 ancestral causal regulatory relationships of a (denser) simulated GRN in terms of ROC (\textbf{top row}), PRC (\textbf{middle row}), and calibration (\textbf{bottom row}) from 100 samples (\textbf{left column}) and 1000 samples (\textbf{right column}). We ran Trigger three times on the simulated data and averaged the results (`trigger') to account for differences when sampling the null statistics. We report the results of three BFCS versions: two described in Algorithm~\ref{alg:BFCS_yeast} for which we take a uniform prior over DAGs (`BFCS DAG') and DMAGs (`BFCS DMAG'), respectively, and one (`BFCS loclink') in which we use the Trigger local-linkage selection strategy described in Subsection~\ref{ssec:finding_causal_links}. For reference, we also show the performance of an equivalent method that uses the BGe score (`BGe').} \label{fig:plot_lG_anc}
\end{figure}

Since BFCS and Trigger are able to pick up ancestral (direct or indirect) causal links, it behooves us to (also) use the transitive closure of the network structure as the gold standard against which to evaluate these methods.  When looking at ancestral causal regulatory relationships, all algorithms show an improvement in calibration (Figure~\ref{fig:plot_sG_anc} compared to Figure~\ref{fig:plot_sG_dir} and Figure~\ref{fig:plot_lG_anc} compared to Figure~\ref{fig:plot_lG_dir}), but also a deterioration in the AUC of ROC and PRC. There are more causal relationships picked up by the inference algorithms that are now labeled as correct, but there are also many more relations to be discovered: 73 ancestral versus 51 direct for the sparser network and 970 ancestral versus 225 direct for the denser network.

\subsection{Comparing Results from an Experiment on Yeast} \label{ssec:yeast}

\begin{algorithm}[!htb]
	\caption{Running BFCS on the yeast data set} 
	\label{alg:BFCS_yeast}
	\begin{algorithmic}[1] 
		\State \textit{Input:} Yeast data set consisting of 3244 markers and 6216 gene expression measurements
		\ForAll{expression traits $T_i$} \label{line:BFCS_yeast:loop1}
			\ForAll{expression traits $T_j, j \neq i$} \label{line:BFCS_yeast:loop2}
				\ForAll{genetic markers $L_k$} \label{line:BFCS_yeast:loop3}
					\State Compute the Bayes factors for the triplet $(L_k, T_i, T_j)$ \label{line:BFCS_yeast:compute_Bf}
					\State Derive the posterior probability of the structure $L_k \rightarrow T_i \rightarrow T_j$ given the data \label{line:BFCS_yeast:derive_post_prob}
				\EndFor
				\State Save $\max_k p(L_k \rightarrow T_i \rightarrow T_j)$ as the probability of gene $i$ regulating gene $j$
			\EndFor
		\EndFor
		\State \textit{Output:} Matrix of regulation probabilities 
	\end{algorithmic}
\end{algorithm}

Chen et al. showcased the Trigger algorithm in~\cite{chen_harnessing_2007} by applying it to an experiment on yeast, in which two distinct strains were crossed to produce 112 independent recombinant segregant lines. Genome-wide genotyping and expression profiling were performed on each segregant line. We computed the probabilities over the triples in the yeast data set using \texttt{trigger}~\citep{chen_trigger_2017} and BFCS (Algorithm~\ref{alg:BFCS_yeast}). For BFCS, we used DMAGs to allow for the possibility of latent variables. 

\begin{table}[!htb]
	\tiny \centering
	\begin{subtable}{.49\linewidth}
		\begin{tabular}{c|cccc}
			Rank & Gene & Chen et al. & trigger & BFCS \\ \hline
			1 & MDM35 & 0.973 & 0.999 & 0.678 \\ 
			2 & CBP6 & 0.968 & 0.997 & 0.683 \\ 
			3 & QRI5 & 0.960 & 0.985 & 0.678 \\ 
			4 & RSM18 & 0.959 & 0.984 & 0.672 \\ 
			5 & RSM7 & 0.953 & 0.977 &  0.684 \\ 
			6 & MRPL11 & 0.924 & 0.999 & 0.670 \\ 
			7 & MRPL25 & 0.887 & 0.908 & 0.675 \\ 
			8 & DLD2 & 0.871 & 0.896 &  0.660 \\ 
			9 & YPR126C & 0.860 & 0.904 &  0.634 \\ 
			10 & MSS116 & 0.849 & 0.997 & 0.659 \\ 
		\end{tabular}
		\subcaption{\tiny Genes regulated by NAM9,  sorted by `Chen et al.'.}
		\label{subtab:NAM9_reg_chen} 
	\end{subtable}
	\begin{subtable}{.49\linewidth}
		\begin{tabular}{c|cccc}
			Rank & Gene & Chen et al. & trigger & BFCS \\ \hline
			1 & FMP39 & 0.176 & 0.401 & 0.691 \\ 
			2 & DIA4 & 0.493 & 0.987 & 0.691 \\ 
			3 & MRP4 & 0.099 & 0.260 & 0.691 \\ 
			4 & MNP1 & 0.473 & 0.999 & 0.691 \\ 
			5 & MRPS18 & 0.527 & 0.974 & 0.690 \\ 
			6 & MTG2 & 0.000 & 0.000 & 0.690 \\ 
			7 & YNL184C & 0.299 & 0.768 & 0.690 \\ 
			8 & YPL073C & 0.535 & 0.993 & 0.690 \\ 
			9 & MBA1 & 0.290 & 0.591 & 0.690 \\ 
			10 & ACN9 & 0.578 & 0.927 & 0.690 \\ 
		\end{tabular}
		\subcaption{\tiny Genes regulated by NAM9, sorted by `BFCS'. }
		\label{subtab:NAM9_reg_bfcs} 
	\end{subtable}

	\caption{ The column `Chen et al.' shows the original results of the Trigger algorithm as reported in~\cite{chen_harnessing_2007}. The `trigger' column contains the probabilities we obtained when running the algorithm from the Bioconductor \texttt{trigger} package ~\citep{chen_trigger_2017} on the entire yeast data set with default parameters. The column `BFCS' contains the output of running Algorithm~\ref{alg:BFCS_yeast} on the yeast data set, for which we took a uniform prior over DMAGs.}
	\label{tab:NAM9_reg} 
\end{table}

In Table~\ref{subtab:NAM9_reg_chen}, we report the top ten genes purported to be regulated by the putative regulator NAM9, sorted according to the probability estimates reported in~\cite{chen_harnessing_2007}. We see that BFCS also assigns relatively high, albeit much more conservative, probabilities to the most significant regulatory relationships found by Trigger. In Table~\ref{subtab:NAM9_reg_bfcs}, we see that some relationships ranked significant by BFCS are assigned a very small probability by Trigger. The regulatory relationship NAM9 $\rightarrow$ MTG2, of which both genes are associated with the mitochondrial ribosome assembly~\cite{the_uniprot_consortium_uniprot_2018}, is ranked sixth by BFCS, but is assigned zero probability by Trigger. This is because the genetic marker $L_i$ exhibiting the strongest linkage with $T_i$ (NAM9 in this case) is preselected in the Trigger algorithm and only the probability of $L_i \rightarrow T_i \rightarrow T_j$ given the data is estimated, while other potentially more relevant genetic markers may be filtered out. With BFCS, on the other hand, we estimated the maximum probability of this structure for all genetic markers (see also subsection~\ref{ssec:finding_causal_links}).

In Subsection~\ref{ssec:prior_cstr}, we have computed an upper bound on the posterior probability $p(L_k \to T_i \to T_j \given \data)$, which is dependent on the chosen prior on causal structures and on the number of samples. For 112 observations (in the yeast data set) and a uniform prior on DMAGs with background knowledge (Table~\ref{tab:no_str_3var}), we obtain a conservative upper bound of $0.6909$. This explains why the BFCS probabilities are tightly packed together in the interval $[0.690, 0.691]$ in \Cref{subtab:NAM9_reg_bfcs}. If we use the prior probabilities derived from a uniform prior on DAGs with background knowledge, as in `BFCS DAG', this upper bound increases to $0.7703$ because there are fewer possible `full' models competing against the `causal' model. Whereas our posterior probability predictions are appropriately conservative for this number of observations, as reflected by the derived upper bound, Trigger often returns overconfident predictions.

\section{Computational and time complexity} \label{sec:complexity}

The simplicity and inherent parallelism of our approach makes it suitable for large-scale causal network inference. In this section, we evaluate the computational and time complexity of our algorithm and compare it against similar approaches. Assuming that the data correlation matrix required for deriving the Bayes factors in~\eqref{eqn:bayes_factors} has been precomputed, then the derivation takes constant time. Deriving the posterior probability of $L_k \to T_i \to T_j$ from the Bayes factors also takes constant time, since we always have to look at a fixed number of eleven covariance structures. The computational complexity in Algorithm~\ref{alg:BFCS_yeast} is then driven by the nested iteration loop over all possible $(L_k, T_i, T_j)$ triples in~\cref{line:BFCS_yeast:loop1,line:BFCS_yeast:loop2,line:BFCS_yeast:loop3}. Assuming that there are $m$ genes in the regulatory network for which we have measured expression levels and $l$ genetic markers, we arrive at a computational complexity of $\bigO{lm^2}$.

Any routine for deriving the posterior probability of $L_k \to T_i \to T_j$ can be plugged inside the iteration loops in Algorithm~\ref{alg:BFCS_yeast}. For instance, in the reference BGe procedure (Subsection \ref{ssec:prior_cstr}), we replace~\cref{line:BFCS_yeast:compute_Bf,line:BFCS_yeast:derive_post_prob} with a function that computes the BGe score of each causal structure on triplets. This involves summing up over a number of local scores for each node, where the complexity of each local computation depends on the number of nodes involved (the node and its parents). Since we are only computing the BGe score over triples, the computational complexity is also of order $\bigO{1}$. However, a single BGe score computation is typically more expensive than computing the Bayes factors, since the latter involves a number of simplifications.

\begin{figure}[H]
\begin{minipage}[t]{.49\linewidth}
		\includegraphics[width=\linewidth]{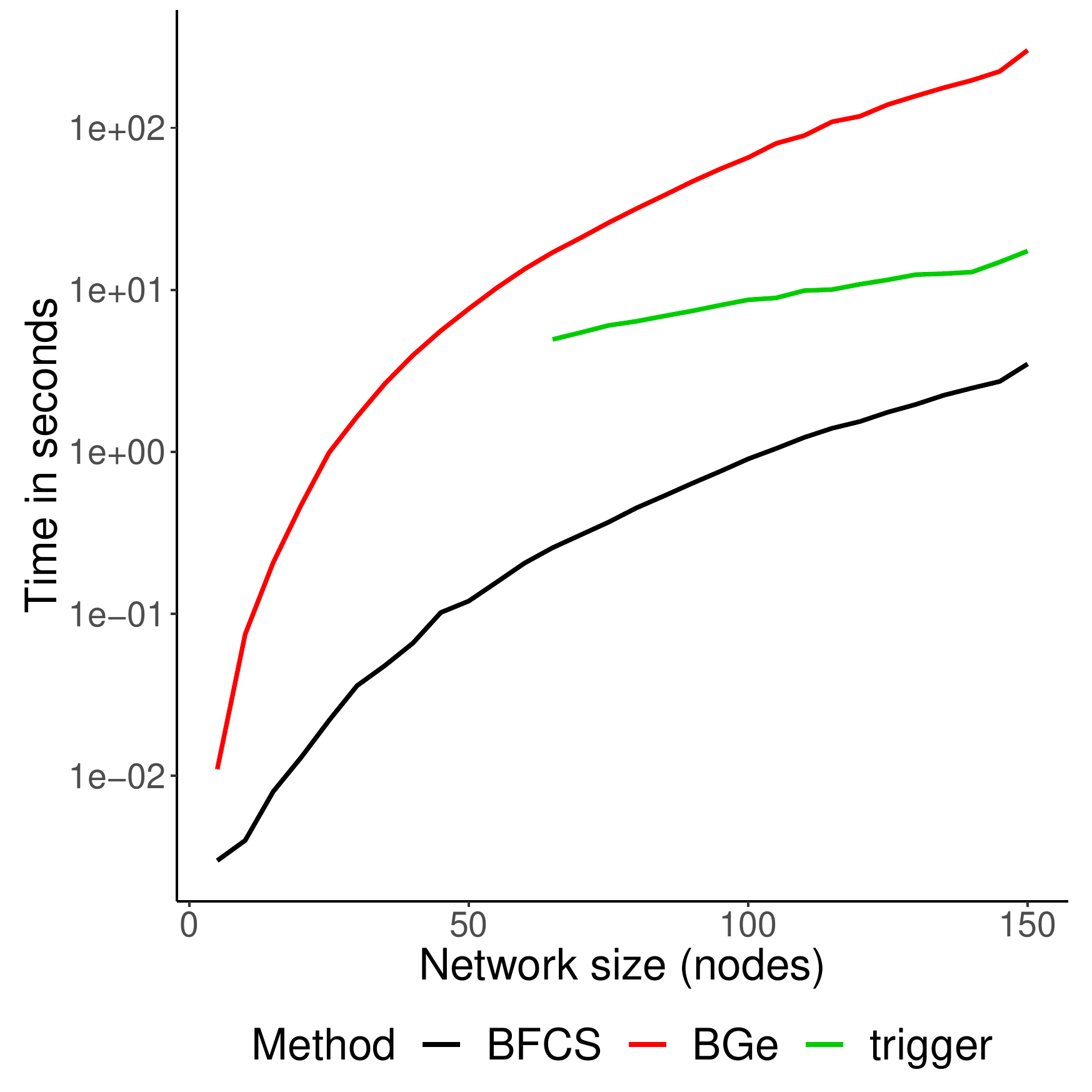}
		\caption{In this simulation, we generated 100 observations from GRNs of increasing network size, where the number of genetic markers was equal to the number of expression traits (nodes). We report the time measurements for BFCS, Trigger, and BGe versus the network size. For networks with fewer than 65 nodes, we ran into problems while running Trigger having to do with the algorithm not being able to produce an estimate of $\pi_0$, the proportion of true null $p$-values. We have observed (not shown here) that the GRN density does not influence the time complexity.  } \label{fig:time_size}
\end{minipage} \hfill %
\begin{minipage}[t]{.49\linewidth}
		\includegraphics[width=\linewidth]{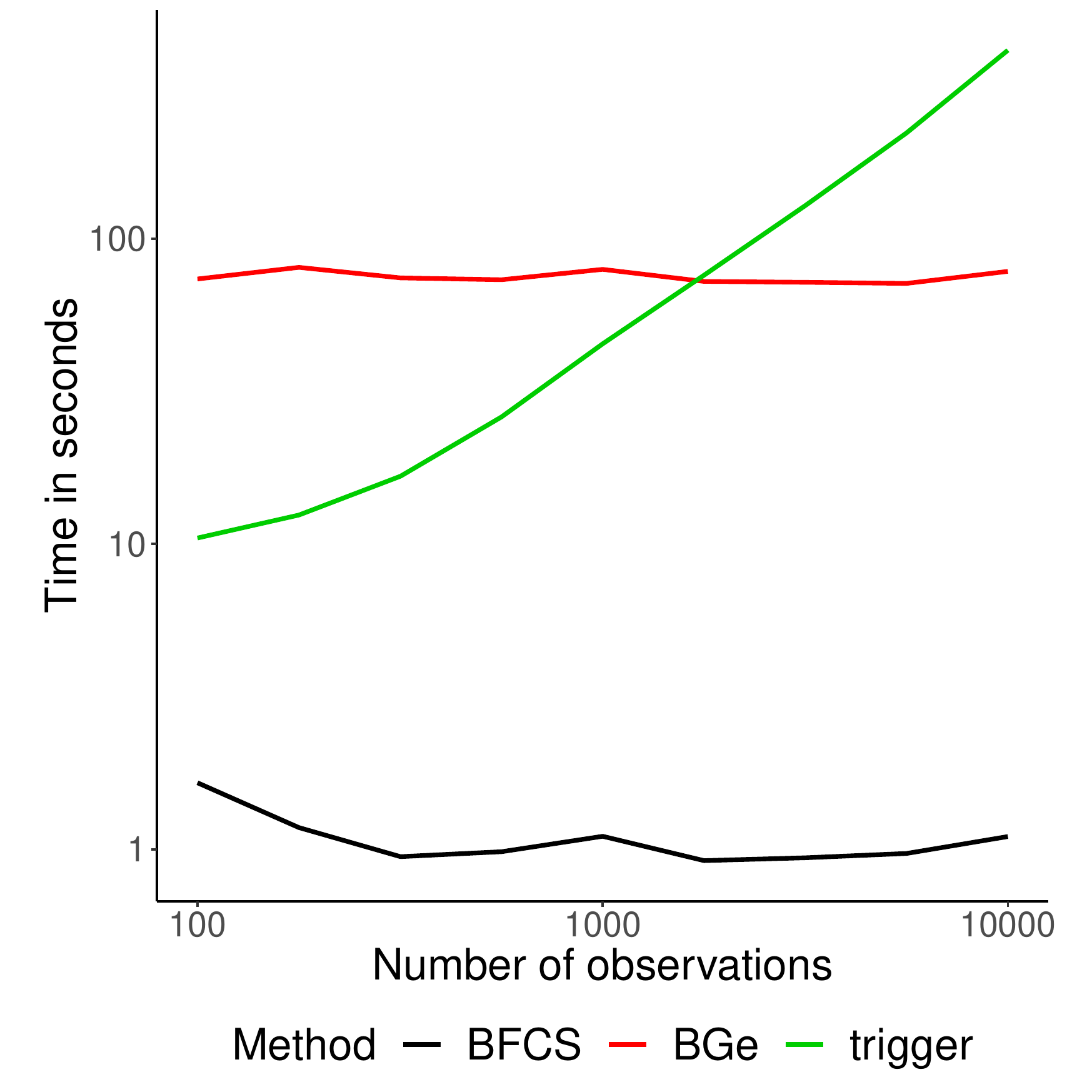}
		\caption{In this simulation, we generated GRNs with 100 expression traits and 100 genetic markers. At each step we increased the number of observations linearly on a logarithmic scale. We report the time measurements in for BFCS, trigger, and BGe versus the number of observations. We see that the time complexity of Trigger is affected by the data sample size, unlike that of BFCS and BGe. For the last two algorithms we included the time it takes to compute the data correlation matrix. } \label{fig:time_nobs}
\end{minipage}
\end{figure}

The computation of the Bayes factors in~\eqref{eqn:bayes_factors} is amenable to vectorization. Instead of passing three correlation coefficients each time to the computation routine, we can pass three vectors of correlation coefficients and perform the computation element-wise. This way we can take advantage of processor architecture supporting SIMD (Single instruction, multiple data) instructions such as \texttt{AVX}~\cite{lomont_introduction_2011}. Since in the vast majority of cases we will have the same number of observations for each triple of variables, we can make further savings by precomputing the prefactors $f(\nobs, \nu)$ and $g(\nobs, \nu)$ in~\eqref{eqn:bayes_factors}.

In~\Cref{fig:time_size,fig:time_nobs} we illustrate the time complexity of the approaches discussed relative to the network size (number of variables) and to the number of data points. The time measurements were taken on an \texttt{HP\textregistered~EliteBook\texttrademark~850 G2} sporting an \texttt{Intel\textregistered~Core\texttrademark~i7-5600U} 2.60 GHz CPU with \texttt{SSE4.1}, \texttt{SSE4.2} and \texttt{AVX2} vectorization capabilities and 16GiB of \texttt{SODIMM DDR3 Synchronous} 1600 MHz RAM. To ensure a fair comparison, we did not include the time it takes Trigger to compute the local linkage between genetic markers and expression traits in the measurement. Figure~\ref{fig:time_size} confirms our expectation that the run time for both BGe and BFCS increases cubically. We notice that BFCS outperforms the other two approaches and is extremely fast even for very large networks. Trigger scales better with the size of the network since it preselects the genetic marker with the strongest linkage for each expression trait pair, essentially reducing the computational complexity from $\bigO{lm^2}$ to $\bigO{m^2}$. However, in Figure~\ref{fig:time_nobs} we see that its execution time is strongly dependent on the amount of samples in the data set. This is because the Trigger algorithm involves computationally expensive operations that require permuting the data set in order to obtain generate null statistics for performing likelihood ratio tests.

\section{Discussion} \label{sec:discussion}

We have introduced a novel Bayesian approach for inferring gene regulatory networks that uses the information in the local covariance structure over triplets of variables to make statements about the presence of causal relationships. The probability estimates produced by BFCS constitute a measure of reliability in the inferred causal relations. One key advantage of our method is that we consider all possible causal structures at once, whereas other methods only look at and test for a subset of structures. We have demonstrated the effectiveness of our approach by comparing it against the Trigger algorithm, a state-of-the-art procedure for inferring causal regulatory relationships. Other methods for inferring gene regulatory networks such as `CIT'~\citep{millstein_disentangling_2009} or `CMST'~\citep{neto_modeling_2013} output $p$-values instead of probability estimates, which is why they are not directly comparable to BFCS.

Since we focus on discovering local causal structures, our method is simple, fast, and inherently parallel, which makes it applicable to very large data sets. Moreover, this enables us to consider more (and possibly better) candidates for estimating the probability of causal regulatory relationships given data than the more restricted search strategy employed by Trigger. Cooper~\cite{cooper_simple_1997}, as well as Silverstein et al.~\cite{silverstein_scalable_2000}, warn that even when the causal graphs contain only a few variables, exact computation is often intractable with Bayesian methods. In our Bayesian approach, the derivation becomes tractable due to the particular model assumptions made. If we are not willing to assume a (latent) Gaussian model, then sampling or approximating methods may be a promising alternative~\cite{cooper_simple_1997}.

In this paper, we have proposed simple uniform priors on two types of causal graph structures, namely DAGs and DMAGs. However, our method allows for more informative causal priors to be incorporated, taking into consideration properties such as the sparsity of the networks. Moreover, our approach is structure-agnostic, by which we mean we can consider different causal graph structures incorporating various data-generating assumptions. The tricky part is then to come up with an appropriate prior on the set of causal graph structures from which we assume the data is generated.

Our approach could be extended to other types of local structures in future work. A straightforward idea would be to consider more than three variables in a local causal structure, which may allow us to discover causal links that cannot be found by looking only at triplets.
 
The code implementing the proposed method will be made publicly available by the authors at \url{https://github.com/igbucur/BFCS}.

\section*{Acknowledgments}

This research has been partially financed by the Netherlands Organisation for Scientific Research (NWO), under project 617.001.451.  We would like to thank the anonymous reviewers for their thoughtful comments and efforts towards improving our manuscript.

\newpage

\appendix

\section{Proofs}

\begin{lemma}[Completeness of covariance structures] There are only eleven distinct non-degenerate (full rank) covariance structures over three variables, which correspond to the five canonical cases described in Figure~\ref{fig:equivalence}.
	
	\label{lem:complete_cov_str}
	
	\begin{proof} 
		
		We have three off-diagonal terms in the covariance matrix $\cov = (\cove_{ij})$  and three off-diagonal terms in the precision matrix $\icov = (\icove_{ij})$ that can be constrained to zero. This means that we have to consider $2^6 = 64$ configurations of zeros. However, a constraint in the covariance matrix implies a constraint in the precision matrix and vice versa:
		
		$$ \icove_{ij} = 0 \implies \cove_{ij} \cove_{kk} = \cove_{ik} \cove_{kj} \quad \land \quad \cove_{ij} = 0 \implies \icove_{ij} \icove_{kk} = \icove_{ik} \icove_{kj},$$
		for any distinct $i, j, k \in \{1, 2, 3\}$. This leads to four implications, whereby if two terms are constrained to zero, (at least) two more have to be constrained to zero:
		
		\begin{enumerate}
			\item $\left\{ \begin{matrix} \icove_{ik} = 0 \\ \icove_{jk} = 0 \end{matrix} \right. \implies \left\{ \begin{matrix} \cove_{jj} \cove_{ik} = \cove_{ij} \cove_{jk} \\ \cove_{ii} \cove_{jk} = \cove_{ij} \cove_{ik} \end{matrix} \right. \implies \left\{ \begin{matrix} \cove_{ik} (\cove_{ii} \cove_{jj} - \cove_{ij}^2) = 0 \\ \cove_{jk} (\cove_{ii} \cove_{jj} - \cove_{ij}^2) = 0 \end{matrix} \right. \overset{\cov \; p.d.}{\implies} \left\{ \begin{matrix} \cove_{ik} = 0 \\ \cove_{jk} = 0 \end{matrix} \right. $
			\item $\left\{ \begin{matrix} \cove_{ik} = 0 \\ \cove_{jk} = 0 \end{matrix} \right. \implies \left\{ \begin{matrix} \icove_{jj} \icove_{ik} = \icove_{ij} \icove_{jk} \\ \icove_{ii} \icove_{jk} = \icove_{ij} \icove_{ik} \end{matrix} \right. \implies \left\{ \begin{matrix} \icove_{ik} (\icove_{ii} \icove_{jj} - \icove_{ij}^2) = 0 \\ \icove_{jk} (\icove_{ii} \icove_{jj} - \icove_{ij}^2) = 0 \end{matrix} \right. \overset{\icov \; p.d.}{\implies} \left\{ \begin{matrix} \icove_{ik} = 0 \\ \icove_{jk} = 0 \end{matrix} \right.$
			\item $\left\{ \begin{matrix} \cove_{ij} = 0 \\ \icove_{ij} = 0 \end{matrix} \right. \implies \left\{ \begin{matrix} \icove_{kk} \icove_{ij} = \icove_{ik} \icove_{jk} = 0 \\ \cove_{kk} \cove_{ij} = \cove_{ik} \cove_{jk} = 0 \end{matrix} \right. \implies \left\{ \begin{matrix} \icove_{ik} = 0 \lor \icove_{jk} = 0 \\ \cove_{ik} = 0 \lor \cove_{jk} = 0 \end{matrix} \right.$
			\item $\left\{ \begin{matrix} \cove_{ik} = 0 \\ \icove_{jk} = 0 \end{matrix} \right. \implies \left\{ \begin{matrix} \icove_{jj} \icove_{ik} = \icove_{ij} \icove_{jk} = 0 \\ \cove_{ii} \cove_{jk} = \cove_{ij} \cove_{ik} = 0 \end{matrix} \right. \overset{\cov, \icov \; p.d.}{\implies} \left\{ \begin{matrix} \icove_{ik} = 0 \\ \cove_{jk} = 0 \end{matrix} \right.$
		\end{enumerate}
		
		Let $z \in \{0, 1, 2, 3\}$ be the number of zeros in the covariance matrix. We explore the covariance structures in increasing order of $z$.
		
		\begin{itemize}
			\item $z = 0:$ 1) If there are also no zeros in the precision matrix, then we have a `full' covariance structure.
			2) If there is a zero in the precision matrix, then we have one of the three `causal' covariance structures. 3) If there are two or more zeros in the precision matrix, then there are also zeros in the covariance matrix because of implication 1. This leads to a contradiction.

			\item $z = 1:$ 1) If there are no zeros in the precision, then we have one of the three `acausal' covariance structures.
			2) If there is at least one zero in the precision matrix, then via implication 3 or implication 4, we also have at least two zeros in the covariance matrix. This leads to a contradiction.

			\item $z = 2$: Because of implication 2, there are also at least two zeros in the precision matrix. 1) If there are only these two zeros in the precision matrix, then we have one of the three `independent' covariance structures. 2) If all off-diagonal terms in the precision matrix are zero, then all three off-diagonal terms in the covariance matrix are zero (implication 1). This leads to a contradiction.

			\item $z = 3$: In this situation, the covariance matrix is diagonal, in which case the precision matrix is also diagonal. This means we have an `empty' covariance structure.
			
		\end{itemize}
		
		In conclusion, we have determined that there are only eleven distinct full rank covariance structures corresponding to the five canonical cases in Figure~\ref{fig:equivalence}: one `full', three `causal', three `acausal', three `independent' and one `empty'.		
	\end{proof}
\end{lemma}

\newpage

\section*{References}

\vskip 0.2in
\bibliography{BFCS_arXiv}

\end{document}